\documentclass[letterpaper, 10 pt, journal, twoside]{IEEEtran}
\usepackage{amsmath,amsfonts}
\usepackage{algorithm}
\usepackage{algpseudocode}
\usepackage{array}
\usepackage{subfigure}
\usepackage{textcomp}
\usepackage{stfloats}
\usepackage{url}
\usepackage{verbatim}
\usepackage{graphicx}
\usepackage{balance}
\usepackage[T1]{fontenc}
\usepackage{dsfont}
\newtheorem{remark}{Remark}
\usepackage{color}
\usepackage[numbers]{natbib}
\usepackage{multicol}
\usepackage[bookmarks=true]{hyperref}
\usepackage{amssymb}
\usepackage{bm}
\usepackage{float}
\usepackage{tikz}
\usepackage{bbm}
\usepackage{booktabs, adjustbox}
\usepackage{setspace}
\DeclareMathOperator*{\argminB}{argmin}

\usepackage{mathtools}
\usepackage{comment}
\DeclarePairedDelimiterX{\infdivx}[2]{(}{)}{%
  #1\;\delimsize\|\;#2%
}

\usepackage{tabularx}
\usepackage{xcolor}

\newcommand{\edit}[1]{\textcolor{black}{#1}}

\algdef{SE}[PFor]{PFor}{EndPFor}[1]{\textbf{parallel for} #1}{\textbf{end parallel for}}

\begin{document}

\title{SIT-LMPC: Safe Information-Theoretic \\Learning Model Predictive Control for Iterative Tasks}

\author{
\thanks{Manuscript Received: June 26, 2025; Revised: September 28, 2025; Accepted: October 24, 2025.}
\thanks{This paper was recommended for publication by
Editor Soon-Jo Chung upon evaluation of the Associate Editor and Reviewers’ comments.}
\thanks{This work was partially supported by US DoT Safety21 National University Transportation Center and NSF grants CISE-$2431569$ and 2514584.}%
Zirui~Zang, Ahmad~Amine, Nick\mbox{-}Marios~T.~Kokolakis,
Truong~X.~Nghiem, Ugo~Rosolia, and Rahul~Mangharam%
\thanks{Z. Zang, A. Amine, N.-M. T. Kokolakis, and R. Mangharam are with the University of Pennsylvania, Philadelphia, PA 19104, USA. E-mail: \{zzang, aminea, nmkoko, rahulm\}@seas.upenn.edu.}%
\thanks{T. X. Nghiem is with the Department of Electrical and Computer Engineering, University of Central Florida, Orlando, FL 32816, USA. E-mail: truong.nghiem@ucf.edu.}%
\thanks{U. Rosolia is with Lyric, Italia 152 Catania Italy. E-mail: ugo.rosolia@gmail.com}%
\thanks{Z. Zang and A. Amine contributed equally to this work.}%
\thanks{Supplementary videos: \href{https://sites.google.com/view/sit-lmpc/}{https://sites.google.com/view/sit-lmpc/}}
\thanks{Digital Object Identifier (DOI): see top of this page.}
}

\markboth{IEEE Robotics and Automation Letters. Preprint Version. Accepted October, 2025}
{Zang \MakeLowercase{\textit{et al.}}: SIT-LMPC: Safe Information-Theoretic Learning Model Predictive Control for Iterative Tasks} 

\maketitle

\begin{abstract}
Robots executing iterative tasks in complex, uncertain environments require control strategies that balance robustness, safety, and high performance. This paper introduces a safe information-theoretic learning model predictive control (SIT-LMPC) algorithm for iterative tasks. Specifically, we design an iterative control framework based on an information-theoretic model predictive control algorithm to address a constrained infinite-horizon optimal control problem for discrete-time nonlinear stochastic systems. An adaptive penalty method is developed to ensure safety while balancing optimality. Trajectories from previous  iterations are utilized to learn a value function using normalizing flows, which enables richer uncertainty modeling compared to Gaussian priors. SIT-LMPC is designed for highly parallel execution on graphics processing units, allowing efficient real-time optimization. Benchmark simulations and hardware experiments demonstrate that SIT-LMPC iteratively improves system performance while robustly satisfying system constraints.
\end{abstract}

 \IEEEpeerreviewmaketitle
\section{Introduction}

\begin{figure*}[t]
    \centering
    \includegraphics[width=\linewidth]{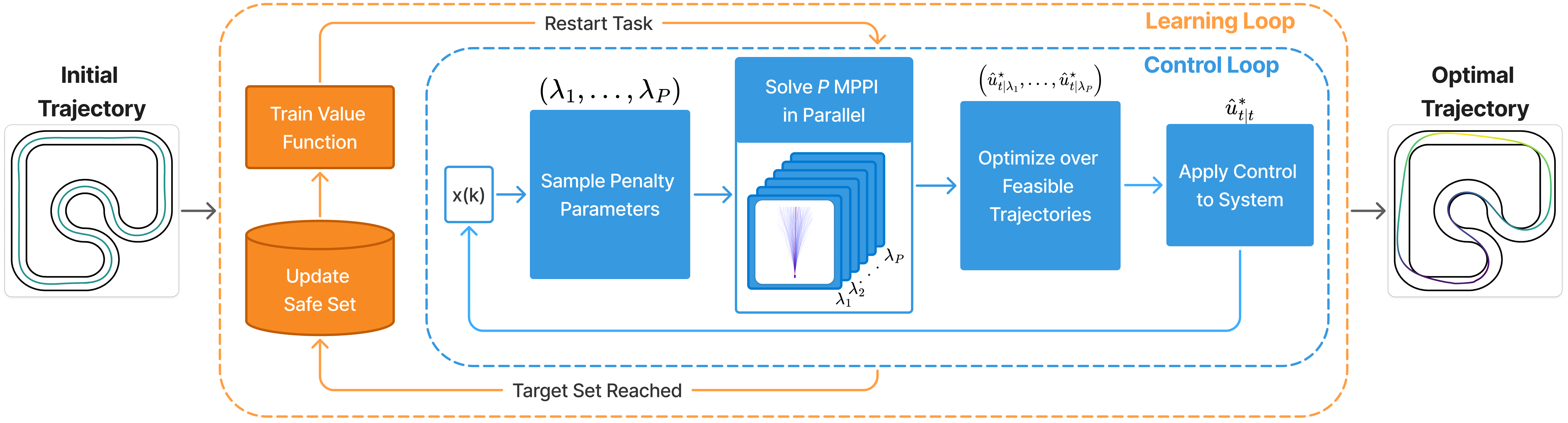}
    \caption{\edit{SIT-LMPC architecture: starting from an initial trajectory, the algorithm iteratively updates the safe set and value function model (orange loop), while solving multiple MPPI problems in parallel (blue loop) to generate optimal trajectories.}}
    \label{img/title}
\vspace{+0.5em}
\end{figure*}

\IEEEPARstart{I}{terative} tasks are ubiquitous in robotics, spanning applications such as quadrotor racing \cite{kaufmann2023champion}, quadruped motion planning \cite{ding2023robust}, and surgical robotics \cite{van2010superhuman}. The core challenge in these settings lies in improving performance over successive executions, leveraging data from prior attempts to refine future behavior \cite{zhao2024survey}. To achieve this, a variety of approaches have been explored, \edit{including deep reinforcement learning (RL) \cite{fishman2023motion}, genetic algorithms \cite{pehlivanoglu2021enhanced}, and optimal control \cite{ding2023robust,schoellig2012optimization}}. These iterative tasks usually involve additional complexities, such as navigating dynamic environments with obstacles \cite{rosolia2021time} or ensuring safe human-robot interaction \cite{goodrich2008human}. Thus, balancing performance with safety through constraint satisfaction during training is a key requirement.

\textit{Iterative learning control} (ILC) improves system performance by using error information from previous task executions to refine future control signals \cite{ILCbristow2006survey}. \textit{Learning model predictive control} (LMPC) \cite{lmpcTAC} is a reference-free variant of ILC that iteratively constructs a controlled invariant terminal constraint set (safe set) and a terminal cost function within a model predictive control (MPC) framework, optimizing for solutions to constrained infinite-horizon optimal control problems over successive iterations. LMPC ensures safety by enforcing state constraints throughout the MPC horizon and enforcing the terminal state to reside within the safe set. The control method converges asymptotically to the optimal controller for deterministic linear systems with quadratic costs \cite{rosolia2022optimality}. For stochastic systems, \cite{rosolia2021robust} specializes LMPC to linear systems with state noise by constructing robust safe sets from previous trajectories and learning a terminal cost function representing the value function associated with the control policies used for collecting data. These safe sets and terminal cost function can be approximated using a finite number of prior trajectories while ensuring that the worst-case iteration cost is non-increasing \cite{samplingLMPC_Ugo}. Recently, in adjustable boundary condition (ABC)-LMPC \cite{thananjeyan2020safety,thananjeyan2021abc}, the LMPC framework is extended to stochastic nonlinear dynamical systems by using a sampling-based cross-entropy method (CEM) MPC to repeatedly sample trajectories until all samples satisfy the terminal set constraint and select the least-cost sampled trajectory \cite{rubinstein2004cross}. Although ABC-LMPC theoretically extends the LMPC framework to stochastic nonlinear dynamical systems, the state constraints are encoded with a high constant cost in the cost function, which yields overly conservative control. Furthermore, it has been reported that the CEM sampling leads to infeasible solutions for stochastic systems and is susceptible to mode collapse in high-dimensional nonlinear systems \cite{williams2017information}.

\textit{Information-theoretic MPC} or \textit{model predictive path integral control} (MPPI) \cite{williams2016aggressive} is a sampling-based MPC algorithm for stochastic systems. MPPI synthesizes the optimal control by minimizing the Kullback-Leibler (KL) divergence between the optimal control distribution and the sampled control distribution \cite{williams2016aggressive}. Stochasticity is handled by sampling trajectories and optimizing over their expected cost without requiring gradient information. The sampling process can be parallelized on a graphics processing unit (GPU), enabling real-time control. Previous work has demonstrated that MPPI outperforms CEM-MPC in terms of safety and cost \cite{williams2018information}. However, MPPI is an unconstrained optimal control method, and state constraints are typically incorporated through clamping in the dynamics or a high cost on violations \cite{williams2018information}. Alternatively, state constraints can be satisfied by projecting unsafe sampled trajectories onto a feasible set for differentially flat systems \cite{rastgar2024priest} or by incorporating a control barrier function (CBF) into the cost function combined with a gradient-based local repair step \cite{yin2023shield}. These approaches can ensure safety, but rely on special assumptions about model dynamics or the existence of a valid CBF, which is hard to realize for real-world data-based scenarios. To the best of our knowledge, there is no constrained MPPI that can handle general state constraints.

The main contribution of this paper is the development of a safe iterative learning control framework for general stochastic nonlinear systems by extending the LMPC formulation and solving the resulting optimization problem by designing a constrained information-theoretic MPC algorithm. Our proposed approach is general and does not rely on assumptions about the system's dynamics and state constraints. To efficiently and effectively balance optimality and safety, we develop an online sampling-based adaptive penalty method. We learn the value function using normalizing flows by leveraging trajectories from previous iterations, enabling richer uncertainty modeling than Gaussian priors. We provide a fully parallelized deployment of our method, enabling $100$Hz+ real-time control on a scaled off-road vehicle with an NVIDIA Jetson Orin AGX. \edit{The architecture of the proposed SIT-LMPC framework is illustrated in Fig. \ref{img/title}}.

\section{Problem Formulation}

In this section, we state the \textit{constrained infinite horizon optimal control problem} for discrete-time nonlinear stochastic systems to characterize an admissible feedback controller that \textit{i}) renders a set of admissible states and a target set robust controlled invariant, \textit{ii}) robustly drives the system state to the target set asymptotically, and \textit{iii}) optimizes the system performance.

Consider the discrete-time nonlinear stochastic dynamical system given by
\begin{align}
\label{equation/dynamics}
    x(k+1) = f\bigl(x(k), u(k), w(k)\bigr), \quad x(0) \stackrel{\text { a.s. }}{=} x_0,
\end{align}
where, for every $ k \in \overline{\mathbb{Z}}_{+} \triangleq \{0, 1,2, \ldots\}$,  $x(k) \in \mathbb{R}^{n_x}$ is a state vector, $u(k) \in \mathbb{R}^{n_u}$ is a control input, $w(k) \in \mathcal{W}\subset \mathbb{R}^{n_w}$ is an independent and identically distributed (i.i.d.) stochastic disturbance process with $\mathcal{W}$ being a bounded set, and $f: \mathbb{R}^{n_x} \times \mathbb{R}^{n_u} \times \mathcal{W} \rightarrow \mathbb{R}^{n_x} $ is jointly continuous in $x,\ u,$ and $w$. The initial condition $x(0)$ is assumed to be a constant vector, almost surely (a.s.) equal to $x_0$.

Let $\mathcal{X} \subset \mathbb{R}^{n_x}$ be a set of \textit{admissible states} and let $\mathcal{U} \subset \mathbb{R}^{n_u}$ be a set of \textit{admissible control inputs}. Furthermore, let $\mathcal{T} \subset \mathcal{X} $ be a set of \textit{target states} and assume that $\mathcal{T}$ is a \textit{robust controlled invariant set} \cite{rungger2017computing} with respect to the discrete-time nonlinear stochastic dynamical system \eqref{equation/dynamics} and the admissible control inputs $\mathcal{U}$, that is, for every $x \in \mathcal{T}$, there exists $u(x) \in \mathcal{U}$ such that $f(x,u(x), w) \in \mathcal{T}$ for all $w \in \mathcal{W}$. In other words, the target set $\mathcal{T}$ is a robust controlled invariant set with respect to \eqref{equation/dynamics} and $\mathcal{U}$ if, for every initial condition $x_0 \in \mathcal{T}$, there exists an admissible feedback control law $u: \mathcal{T} \rightarrow \mathcal{U}$ such that the solution sequence $x(k),\ k \in \overline{\mathbb{Z}}_{+},$ to \eqref{equation/dynamics} remains a.s. in $\mathcal{T}$ for all disturbance sequences $w(\cdot)$. A set $\mathcal{S}\subset \mathcal{X}$ is said to be \textit{safe} if it is a robust controlled invariant set with respect to \eqref{equation/dynamics} and $\mathcal{U}$. Hence, $\mathcal{T}$ is safe.

To evaluate the performance of the discrete-time nonlinear stochastic dynamical system \eqref{equation/dynamics} over the infinite horizon, we define, for every $x_0 \in \mathbb{R}^{n_x}$ and control input $u(k),\ k \in \overline{\mathbb{Z}}_{+}$, the performance measure
\begin{equation} \label{performance measure}
    J\bigl(x_0,u(\cdot) \bigr) \triangleq {\mathbb{E}} \left[ \sum_{k=0}^{\infty} h\bigl(x(k),u(k)\bigr) \right],
\end{equation}
where $h: \mathbb{R}^{n_x} \times \mathbb{R}^{n_u}\rightarrow \mathbb{R} $ is the stage cost satisfying $h(x,u)=0$ for every $(x,u) \in \mathcal{T} \times \mathbb{R}^{n_u}$ and $h(x,u)>0$ for every $(x,u) \in \left(\mathbb{R}^{n_x}  \backslash \mathcal{T} \right) \times \mathbb{R}^{n_u}$.  That is, the target set $\mathcal{T}$ incurs zero cost.

The \edit{goal} is the synthesis of an admissible feedback control law $u^\star: \mathcal{X} \rightarrow \mathcal{U}$ that renders $\mathcal{X}$ and $\mathcal{T}$ robust controlled invariant, robustly drives, for every $x_0 \in \mathcal{X}$, 
the state $x(k),\ k \in \overline{\mathbb{Z}}_{+},$ 
to the target set $\mathcal{T}$ 
as $k \rightarrow \infty$, and minimizes the performance measure \eqref{performance measure}. 

In light of the above, our control problem can be cast as a \textit{constrained infinite horizon optimal control problem} given by
\begin{subequations}
\label{equation/infiniteOCP}
\begin{align}
    J^\star(x_0) \triangleq  & \min_{u(\cdot) \in \mathcal{F}(x_0)}  J\bigl(x_0,u(\cdot)\bigr)
    \label{equation/problem0} \\
    \textrm{subject to} \nonumber \\
    & x(k+1) = f\bigl(x(k), u(k), w(k)\bigr), \quad k \in \overline{\mathbb{Z}}_{+},\\
    & x(0) \stackrel{\text { a.s. }}{=}  x_0 \in \mathcal{X},\\
    &x(k) \in \mathcal{X},\quad u(k) \in \mathcal{U},\quad  k \in \overline{\mathbb{Z}}_{+}, \label{equation/infiniteOCP:constraints}
\end{align}
\end{subequations}
where $\mathcal{F}(x_0)$ denotes the set of \textit{admissible feedback stabilizing controllers}, which is assumed to be nonempty and defined as
\begin{multline*}
    \mathcal{F}(x_0) \triangleq  \bigl\{ u: \mathcal{X} \rightarrow \mathcal{U}: \text{$x(\cdot)$ given by \eqref{equation/dynamics} satisfies a.s. }  \\ 
    x(k) \in \mathcal{X},\ k \in \overline{\mathbb{Z}}_{+},\ \lim _{k \rightarrow \infty} x (k) \in \mathcal{T},\ \text{and}
    \ J\bigl(x_0,u(\cdot)\bigr)< \infty \bigr\}.
\end{multline*}

\section{Safe Information-Theoretic Learning Model Predictive Control}
The constrained infinite-horizon optimal control problem
\eqref{equation/infiniteOCP} involves an optimization over infinite-dimensional function spaces, and hence, it is challenging to solve. In this section, we build on LMPC \cite{lmpcTAC}, MPPI \cite{williams2018information}, and penalty methods for constrained optimization \cite{freund2004penalty} to develop the SIT-LMPC algorithm to address this problem. Specifically, we learn the solution to problem \eqref{equation/infiniteOCP} by iteratively executing the constrained regulation task from the initial condition $ x_0$. System trajectories associated with feasible iterations are stored and used to iteratively synthesize a sampling-based predictive control policy and learn a value function. Constraints are enforced by developing an adaptive penalty method while balancing optimality. 
\subsection{Stochastic LMPC Formulation}
 For every time step $ k \in \overline{\mathbb{Z}}_{+}$, let $x^{l}(k)$, $u^{l}(k)$, and $w^{l}(k)$ be the system state, the control input, and the stochastic disturbance at the $l$th iteration. We assume that at every $l$th iteration, the closed-loop system trajectory starts a.s. from the initial state $ x_0 \in \mathcal{X}$, namely, $x^{l}(0) \stackrel{\text { a.s. }}{=}  x_0, \ l \in \mathbb{Z}_{+}$. An $l$th iteration is said to be \textit{feasible} if 
$ x^{l}(k) \in \mathcal{X}~\mathrm{and}~u^{l}(k) \in \mathcal{U}, \ k \in \overline{\mathbb{Z}}_{+}$, and $\lim _{k \rightarrow \infty} x^{l}(k) \in \mathcal{T}$.
In other words, the $l$th iteration is feasible if the system trajectory $x^{l}(\cdot)$ satisfies the state constraints and converges asymptotically to the target set $\mathcal{T}$ while $x^{l}(\cdot)$ is generated by an admissible control input $u^{l}(\cdot)$. 

Let $\mathcal{I}^l\triangleq \{ i \in[0,\ l]: i \textrm{th iteration is feasible}\}$ be the set of indices of feasible iterations up to the $l$th iteration. Define the \textit{sampled safe set} $\mathcal{S}^l$ at the $l$th iteration as
\begin{equation}
\mathcal{ S}^{l} \triangleq \left\{x^{i}(k) :  
      i \in \mathcal{I}^{l} \ \mathrm{and} \ k \in \overline{\mathbb{Z}}_{+}  \right\} ,
\end{equation}
which is a collection of all states along the system trajectories $x^{i}(\cdot),\ i \in \mathcal{I}^l,$ of feasible iterations up to the $l$th iteration. 
\edit{Following \cite{samplingLMPC_Ugo}, we define $\mathcal{CS}^l$ as the convex hull of $\mathcal{S}^l$}, which is used as a \textit{terminal constraint set} in our framework.

For every iteration $l \in \mathbb{Z}_{+}$, the \edit{$l$th \textit{iteration cost} is defined by $J^{l}(x_0,u^{l}(\cdot)) \triangleq  \sum_{k=0}^{\infty} h(x^{l}(k),u^{l}(k))$, which evaluates the performance of the controller $ u^{l}(\cdot)$. Note that $J^{l}(x_0,u^{l}(\cdot))$ is a random variable depending on the realization of the stochastic disturbance $w(\cdot)$.} The $l$th iteration \textit{value function} $V^{l}(\cdot)$ is defined by $V^l(x_0) \triangleq \mathbb{E}[J^{l}(x_0,u^{l}(\cdot))], \ x_0 \in \mathcal{X},$ and serves as a \textit{terminal cost function} in our SIT-LMPC algorithm.

Now, consider the \textit{constrained finite-time optimal control problem}, for every iteration $l \in \mathbb{Z}_{+}$ and time $t \in \overline{\mathbb{Z}}_{+}$, given the terminal cost function $ V^{l}(\cdot)$, the terminal constraint set $\mathcal{S}^{l-1}$, and a prediction horizon $T \in \mathbb{Z}_{+}$,
\begin{subequations}
\label{CFTOCP}
\begin{align}
& J_\text{LMPC}^{l}\left( x^{l}(t)\right) \nonumber\\ 
&\triangleq \min _{u^{l}_{\cdot \mid t}} {\mathbb{E}} \left[\sum_{k=t}^{t+T-1} h(x^{l}_{k \mid t}, u^{l}_{k \mid t})+V^{l}(x^{l}_{t+T \mid t})\right] \label{equation/problem1} \\
& \text {subject to} \nonumber \\
& x^{l}_{k+1 \mid t}=f\bigl(x^{l}_{k \mid t}, u^{l}_{k \mid t},w^{l}(k)\bigr), \quad k \in\{t, \ldots, t+T-1\} \\
& x^{l}_{t \mid t} \stackrel{\text { a.s. }}{=}x^{l}(t), \\
& x^{l}_{k \mid t} \in \mathcal{X},\quad u^{l}_{k \mid t} \in \mathcal{U} \quad \text{a.s.}, \quad  k \in\{t, \ldots, t+T-1\}, \label{equation/constraint1} \\
& x^{l}_{t+T \mid t} \in \mathcal{CS}^{l-1} \quad \text{a.s.},  \label{equation/constraint2}
\end{align}
\end{subequations}
where, for every $ k \in \{t, \ldots, t+T\}$, $x^{l}_{k \mid t}$ and $u^{l}_{k \mid t}$ denote the predicted value of state $x^{l}(k)$ and input $u^{l}(k)$ given a measurement $x^{l}(t)$, and  \edit{$u^{l}_{\cdot \mid t}$ denotes the predicted control sequence  defined by
$u^{l}_{\cdot \mid t} \triangleq [u^{l}_{t \mid t}, u^{l}_{t+1 \mid t}, \ldots, u^{l}_{t+T-1 \mid t} ]$}.

\begin{remark}
The initial sampled safe set $ \mathcal{S}^0$ is assumed to be nonempty and contains a collection of admissible state trajectories, generated by admissible controllers, that asymptotically converge to $\mathcal{T}$. For instance, $\mathcal{S}^0$ can be constructed from safe human demonstrations or low-performance safe controllers.
\end{remark}

\subsection{Safe MPPI via an Online Adaptive Penalty Method}
\label{subsection:AP-MPPI}
Since MPPI does not consider state constraints, MPPI cannot be used to solve the constrained finite-time optimal control problem \eqref{CFTOCP}. \edit{To address this limitation, we convert the constrained finite-time optimal control problem \eqref{CFTOCP} into an \textit{unconstrained finite-time optimal control problem}. To measure the violation of state constraints for every $x \in \mathbb{R}^{n_x}$, we integrate the \textit{exterior penalty functions} $d_{\mathcal{X}}(\cdot)$ and $d_{\mathcal{CS}^{l-1}}(\cdot),\ l \in \mathbb{Z}_{+},$ associated with the set of admissible states $\mathcal{X}$ and the safe set $\mathcal{CS}^{l-1}$ into the stage cost $h(\cdot,\cdot)$ and the terminal cost $ V^{l}(\cdot)$, respectively.}

Define the distance of a point $x \in \mathbb{R}^{n_x}$ to a closed set $C \subseteq \mathbb{R}^{n_x}$ in the Euclidean norm $\|\cdot\|_2$ as $ \operatorname{dist}(x, C)\triangleq\inf_{y \in C} \left\{\left\|x-y\right\|_2 \right\}$. The exterior penalty functions are defined as $d_{\mathcal{X}}(x)\triangleq \operatorname{dist}(x, \mathcal{X})$ and $d_{\mathcal{CS}^{l-1}}(x)\triangleq \operatorname{dist}(x, \mathcal{CS}^{l-1})$ with associated non-negative \textit{penalty parameters} $\lambda_\mathcal{X}$ and $\lambda_{\mathcal{CS}^{l-1}}$. For every iteration $l \in \mathbb{Z}_{+}$ and time $t \in \overline{\mathbb{Z}}_{+}$, the \textit{unconstrained finite-time optimal control problem} is given by
\begin{subequations}
\label{SIT-LMPC}
\begin{align}
& J_\text{SIT-LMPC}^{l}\left( x^{l}(t)\right) \nonumber \\
& \triangleq  \min_{u^{l}_{\cdot \mid t}} {\mathbb{E}} \Biggl[ \sum_{k=t}^{t+T-1} \left( h(x^{l}_{k \mid t}, u^{l}_{k \mid t})+\lambda_\mathcal{X} d_{\mathcal{X}}(x^{l}_{k \mid t})\right) \nonumber \\
&\hspace{1.5cm} +V^{l}(x^{l}_{t+T \mid t})+\lambda_{\mathcal{CS}^{l-1}}d_{\mathcal{CS}^{l-1}}(x^{l}_{t+T \mid t})\Biggr]  \label{equation/problem2} \\
& \text {subject to} \nonumber \\
& x^{l}_{k+1 \mid t}=f\bigl(x^{l}_{k \mid t}, u^{l}_{k \mid t},w^{l}(k)\bigr), \quad k \in\{t, \ldots, t+T-1\}, \\
& x^{l}_{t \mid t} \stackrel{\text { a.s. }}{=} x^{l}(t), \\
&  u^{l}_{k \mid t} \in \mathcal{U} \quad \text{a.s.},\quad \quad  k \in\{t, \ldots, t+T-1\}. \label{equation/control_constraint}  
\end{align}
\end{subequations}
We use MPPI \cite{williams2016aggressive} to solve the stochastic optimal control problem \eqref{SIT-LMPC} from an information-theoretic perspective by minimizing the KL divergence between the optimal predicted control distribution $\mathbb{Q}^{\star l}$ and the open-loop control distribution $\mathbb{Q}^{ l}$. Hence, the optimal control is given by
\begin{equation}
    u^{\star l}_{k \mid t} \triangleq \argminB_{u^{l}_{\cdot \mid t}} D_{\text{KL}}\infdivx{\mathbb{Q}^{\star l}}{\mathbb{Q}^{ l}}, \quad  k \in\{t, \ldots, t+T-1\}.
\end{equation} 
To optimize this KL divergence, we first sample $N$ trajectories $x^{l_s}_{\cdot \mid t},\ s=1,\ldots, N,$ starting from $x^{l}(t)$, which are generated by sampling control sequences $u^{l_s}_{\cdot \mid t},\ s=1,\ldots, N,$ from a \textit{truncated} normal distribution $\mathcal{N}_{\text{trunc}}(u^{\star l}_{\cdot \mid t-1}, \Sigma, \mathcal{U}),$ where $u^{\star l}_{\cdot \mid t-1}$ is the mean value of the \textit{parent} normal distribution, $\Sigma$ is the user-prescribed covariance of the \textit{parent} normal distribution, and $\mathcal{U}$ specifies the truncation range so that the input constraints \eqref{equation/control_constraint} are satisfied \cite{burkardt2014truncated}. In practice, sampling of the control sequences $u^{l_s}_{\cdot \mid t}$ and simulation of the $N$ trajectories generated by these control inputs are executed in parallel on a GPU. Then, we perform \textit{ importance sampling} on $u^{l}_{\cdot \mid t}$ to obtain the approximate optimal predicted control sequence  as
\begin{equation} 
     \hat{u}^{\star l}_{k \mid t} = \sum_{s=1}^N w_s u^{l_s}_{k \mid t}, \quad  k \in\{t, \ldots, t+T-1\},
     \label{eq:importance_sampling_1}
\end{equation}
with importance sampling weights $w_s$ given by
\begin{equation}
    w_s = \frac{\exp\left(-\frac{1}{\tau}J_\text{sampled}^{l}\bigl( x^{l}(t), x^{l_s}_{\cdot \mid t}, u^{l_s}_{\cdot \mid t}\bigr)\right)}
    {\sum_{s=1}^N \exp\left(-\frac{1}{\tau}J_\text{sampled}^{l}\bigl( x^{l}(t), x^{l_s}_{\cdot \mid t}, u^{l_s}_{\cdot \mid t}\bigr)\right)},
     \label{eq:importance_sampling_2}
\end{equation}
where $s=1,\ldots, N$, $\tau>0$ is a temperature parameter that tunes the sharpness of importance sampling, and $ J_\text{sampled}^{l}\bigl( x^{l}(t),x^{l_s}_{\cdot \mid t}, u^{l_s}_{\cdot \mid t}\bigr)$ is a sampled cost function defined as 
\begin{align}
 & J_\text{sampled}^{l}\bigl( x^{l}(t), x^{l_s}_{\cdot \mid t}, u^{l_s}_{\cdot \mid t}\bigr) \nonumber \\
 & \triangleq  \sum_{k=t}^{t+T-1} \Bigl( h(x^{l_s}_{k \mid t}, u^{l_s}_{k \mid t})+\lambda_\mathcal{X} d_{\mathcal{X}}(x^{l_s}_{k \mid t})\Bigr) + V^{l}(x^{l_s}_{t+T \mid t})\nonumber \\ 
 & \hspace{0.4 cm} +\lambda_{\mathcal{CS}^{l-1}}d_{\mathcal{CS}^{l-1}}(x^{l_s}_{t+T \mid t}),\quad s=1,\ldots, N.
 \label{eq:sampled_cost_function}
\end{align}

However, assuming that the penalty parameters $\lambda_\mathcal{X}$ and $\lambda_{\mathcal{CS}^{l-1}}$ \edit{are arbitrary large} constants may yield conservative system behavior, where safety dominates performance. Hence, we develop an online sampling-based \textit{adaptive penalty} (AP) method that generates the penalty parameters $\lambda_\mathcal{X}$ and $\lambda_{\mathcal{CS}^{l-1}}$ in real-time, allowing for a balance between \textit{optimality} and \textit{safety}. Specifically, we sample the penalty parameters $\lambda_\mathcal{X}$ and $\lambda_{\mathcal{CS}^{l-1}}$ \edit{from uniform distributions, that is,
$
\lambda_\mathcal{X} \sim \mathrm{Unif}\bigl(0,\lambda^\mathrm{max}_{\mathcal{X}}\bigr)$ and $
\lambda_{\mathcal{CS}^{l-1}} \sim \mathrm{Unif}\bigl(0,\lambda^\mathrm{max}_{\mathcal{CS}^{l-1}}\bigr),
$
with $\lambda^\mathrm{max}_{\mathcal{X}},\ \lambda^\mathrm{max}_{\mathcal{CS}^{l-1}}>0$. Let $\lambda \triangleq \left[ \lambda_\mathcal{X},\ \lambda_{\mathcal{CS}^{l-1}} \right]^\mathrm{T}$ be the augmented penalty parameter vector and let $\Lambda \subset \Omega \triangleq[0, \lambda^\mathrm{max}_{\mathcal{X}}] \times [0, \lambda^\mathrm{max}_{\mathcal{CS}^{l-1}}]$ be a finite set of samples drawn from the joint uniform distribution.
For every $\lambda \in \Lambda$, we solve the optimal control problem \eqref{SIT-LMPC} whose approximate solution \eqref{eq:importance_sampling_1} is now parametrized by $\lambda$ and written as $\hat{u}^{\star l}_{\cdot \mid t,\lambda}$.
We then construct the set of penalty parameters that generate feasible trajectories as
\begin{align*}
\mathcal{F}^l_t = \bigl\{ 
\lambda \in \Lambda : & \
\hat{x}^{\star l}_{k \mid t, \lambda} \in \mathcal{X},\ k \in\{t, \ldots, t+T\},\\&
\hat{x}^{\star l}_{t+T \mid t, \lambda} \in \mathcal{CS}^{l-1} \bigr\}, \quad l \in \mathbb{Z}_{+},\quad t \in \overline{\mathbb{Z}}_{+}.
\end{align*}
The optimal penalty parameter $\lambda^\star$ is given by}
\edit{
\begin{equation}
\label{eq:control_selection}
\lambda^\star =
\begin{dcases}
  \begin{aligned}[t] 
     & \argminB_{\lambda \in \mathcal{F}^l_t} 
     J^{l}_\text{sampled} \bigl( x^{l}(t), \hat{x}^{\star l}_{\cdot \mid t,\lambda}, \hat{u}^{\star l}_{\cdot \mid t,\lambda}\bigr), 
  \end{aligned}  \text{if } \mathcal{F}^l_t \neq \varnothing, \\
  \begin{aligned}[t]
     & \argminB_{\lambda \in \Lambda} \left( \sum_{k=t}^{t+T-1} d_{\mathcal{X}}(\hat{x}^{\star l}_{k \mid t, \lambda}) + d_{\mathcal{CS}^{l-1}}(\hat{x}^{\star l}_{t+T \mid t, \lambda})\right),
  \end{aligned} \\ \hspace{5 cm} \text{if } \mathcal{F}^l_t = \varnothing.
\end{dcases}
\end{equation}
In other words, the optimal penalty parameter is chosen to minimize the sampled cost $J^{l}_\text{sampled}$ over the set of feasible parameters, or, if this set is empty, to minimize the constraint violation. Note that sampling the penalty parameters $\lambda$ and executing MPPI are performed in parallel, ensuring efficient computation.}

\edit{Now, we apply the first component $\hat{u}^{\star l}_{t \mid t, \lambda^{\star}}$ of the approximate optimal predicted control sequence $\hat{u}^{\star l}_{\cdot \mid t,\lambda^{\star}}$ to the system \eqref{equation/dynamics}, that is, $u^{l}(t)=\hat{u}^{\star l}_{t \mid t,\lambda^{\star}}$.} Next, given the new observed state $x^{l}(t+1)$, the unconstrained finite-time optimal control problem \eqref{SIT-LMPC} is solved at time $t+1$ with $x^{l}_{t+1 \mid t+1} \stackrel{\text { a.s. }}{=} x^{l}(t+1)$, yielding a receding-horizon control strategy.

\begin{remark}
    Interior penalty functions, such as log-barrier functions, would yield infinite costs for any trajectory with a state that violates the constraints, dominating the importance sampling.
\end{remark}

\section{Practical Implementation}
This section outlines two key challenges associated with implementing SIT-LMPC on a real-world platform. 
\subsection{Iteration Value Function Estimation using Normalizing Flows}
\label{section/learning}

Implementing SIT-LMPC requires approximating the $l$th iteration value function $V^l(x), \ x \in \mathcal{X},$ since a closed-form expression for $V^l(\cdot)$ is generally intractable. To this end, we use \textit{normalizing flows} (NFs) to model the distribution of the $l$th iteration cost $J^{l}(x,u^{l}(\cdot)),\ x \in \mathcal{X},$ from the collected data. NFs \cite{papamakarios2021normalizing} are a class of \textit{generative models} that transform a simple latent probability distribution $z$ into a complex target distribution via a sequence of invertible and differentiable mappings $g_{\theta}$ parameterized by $\theta$. NFs are suitable for stochastic systems since they directly learn complex probability distributions, rather than relying on deterministic estimates or restrictive assumptions, such as Gaussian priors. \edit{We choose NFs over other probabilistic modeling methods due to their richer expressiveness \cite{papamakarios2021normalizing} and computational efficiency for real-time deployment.}

We use \textit{neural spline flows} (NSFs) \cite{durkan2019neural} to learn the conditional distribution of $J^{l}(x,u^{l}(\cdot))$ given a state $x \in \mathcal{X}$. \edit{In our implementation, we use an NSF with eight spline segments and four flow layers, each of which has two fully-connected layers with 96 hidden states.} For each iteration $l \in \mathbb{Z}_{+}$, \edit{we train the NSF for 300 epochs with a learning rate of $10^{-4}$} on the dataset 
 \begin{align}
 \label{equation/dataset}
    \hspace{-0.2 cm} \mathcal{D}^{l-1} \triangleq \Bigl\{\Bigl(x^{i}(k), J^{i} \left( x^{i}(k), u^{i}(\cdot)\right) \Bigr) : i \in \mathcal{I}^{l-1},\ k \in \overline{\mathbb{Z}}_{+}\Bigr\},
 \end{align}
which is initialized with $\mathcal{S}^0$ and \edit{subsequently built with system} states and corresponding iteration costs collected from feasible iterations up to the $l-1$th iteration. Specifically, the estimated conditional distribution of $J^{l}(x,u^{l}(\cdot))$ is denoted by $\hat{J}^{l}$ and given by $\hat{J}^{l} = g^l_{\theta}(z, x),$ where $g_{\theta}$ is an \edit{invertible transformation}, $z$ is a sample from the latent distribution $\mathcal{N}(0, 1)$, and $x$ is the system state used as context input. We train our model using stochastic gradient descent by minimizing a negative log-likelihood loss function given by 
\begin{align*}
   \mathcal{L}^l(\theta) = -\sum_{ \mathcal{D}^{l-1} } \log \left( \frac{1}{\sqrt{2\pi}} e^{-\frac{\left( {g^l_{\theta}}^{-1}(\hat{J}^{l}, x)\right)^2}{2}}\left| \frac{\partial {g^l_{\theta}}^{-1}(\hat{J}^{l}, x)}{\partial \hat{J}^{l}} \right| \right) ,
\end{align*}
where ${g^l_{\theta}}^{-1}$ is the inverse function of $g^l_{\theta}$. When performing inference on  $\hat{J}^{l}$ for an arbitrary state $x \in \mathcal{X}$, we first sample a batch of latent variables $z$ from the latent distribution, then evaluate the NSF conditioned on $x$, and finally compute \edit{the iteration value function} estimate as the \edit{\textit{expected value}} of the iteration cost function estimates, that is,  $V_{\theta}^{l}(x) = \underset{z}{\mathbb{E}} \left[ g^l_{\theta}(z,x)\right]$.
\edit{As shown in the ablation study in Section \ref{section/dynamicST}, our NSF method for modeling the conditional distribution of the iteration cost function outperforms the Bayesian neural network (BNN) approach used in \cite{thananjeyan2021abc}.}

\edit{Algorithm~\ref{alg_learning} outlines our iterative learning procedure. At each iteration, the system executes a trajectory until either the target set is reached or feasibility is lost in steps \ref{line:target} and \ref{line:feasibility}, respectively. The approximate optimal control is computed in step \ref{line:control} using Algorithm~\ref{alg_optimization}, which is the AP-MPPI scheme detailed in Section \ref{subsection:AP-MPPI}. The trajectory states and the associated cost-to-go are collected into $\mathcal{D}^l$ in step \ref{line:dataset1}, which is then used to update the safe set $\mathcal{S}^{l}$ and train the NF iteration cost function model $g_\theta^{l+1}$.}

\begin{algorithm}
\caption{SIT-LMPC}
\label{alg_learning}
\begin{algorithmic}[1]
  \Require $N$: number of trajectory samples, $T$: control horizon, $P$: number of $\lambda$ samples, $L$: maximum iterations, $\mathcal{S}^{0}$: initial safe-set.
  \State Initialize $g_\theta^{1}$ with random weights
  \State Sample $\Lambda\leftarrow\{\lambda_i\}$ for $i = 1,\dots,P$ with $\lambda_i \sim \text{Unif}(\Omega)$
  \For{$l = 1,\dots,L$} \label{line:iterations}
  \While{$x^{l}(t) \notin \mathcal{T}$}  \label{line:target}\Comment{\textit{run until target set is reached}}
    \If{$x^{l}(t)\notin\mathcal{X}$} \Comment{\textit{check for constraint violation}}\label{line:feasibility}
      \State \textbf{terminate}
    \EndIf
    \State $\hat{u}^{\star l}_{t \mid t,\lambda^{\star}} \leftarrow $\hyperref[alg_optimization]{AP-MPPI}$\left(x^{l}(t), \mathcal{S}^{l-1}, g_\theta^l,\Lambda, P,  N, T\right)$\label{line:control}
    \State Apply $\hat{u}^{\star l}_{t \mid t,\lambda^{\star}}$ to system \eqref{equation/dynamics}
  \EndWhile
  \If{$\bigcup_{t=0}^\infty\left\{{x^l(t)}\right\}  \subset\mathcal{X}$} \Comment{\textit{if iteration is feasible}}
    \State $\mathcal{S}^{l} \leftarrow \mathcal{S}^{l-1}\cup \left(\bigcup_{t=0}^\infty\left\{{x^l(t)}\right\}\right)$
    \State Build $\mathcal{D}^{l}$ using \eqref{equation/dataset}
  \EndIf \label{line:dataset1}
  \State Train $g^{l+1}_\theta$ on $\mathcal{D}^{l}$ \Comment{\textit{NF from Section \ref{section/learning}}}
  \EndFor
\end{algorithmic}
\end{algorithm}

\subsection{Parallelizing over Control and Penalty Parameter Sampling}

SIT-LMPC is tailored to use parallelized computation to achieve low-latency control. Algorithm \ref{alg_optimization} implements the SIT-LMPC control loop. Solving the optimal control problem \eqref{equation/problem2} involves two layers of sampling: control sequences for MPPI and penalty parameters for the \textit{online adaptive penalty method}. First, we sample control inputs and generate state trajectories in step \ref{line:traj_sample0}-\ref{line:traj_sample1}. Then, we evaluate the cost function \eqref{eq:sampled_cost_function} for every sampled penalty parameter pair $(\lambda_{\mathcal{CS}^l}, \lambda_{\mathcal{X}})$ in step \ref{line:cost} and perform the importance sampling in step \ref{line:importance}. 
To reduce computation, the sampled trajectories $x^{l_s}_{\cdot \mid t}$ are shared between the importance sampling processes associated with different penalty parameters. In step \ref{line:selection}, we select the optimal solution as detailed in \eqref{eq:control_selection}. With GPU parallelization, the latency of the SIT-LMPC is close to solving one MPPI process.

\begin{algorithm}
\caption{AP-MPPI$\left(x^{l}(t),\mathcal{S}^{l-1}, g_\theta^l,\Lambda, P, N, T\right)$}
\label{alg_optimization}
\begin{algorithmic}[1]
  \PFor{$s = 1,\dots,N$} \label{line:traj_sample0}
      \State $u^{l_s}_{\cdot \mid t}\sim \mathcal{N}_{\text{trunc}}(u^{\star l}_{\cdot \mid t-1}, \Sigma, \mathcal{U})$\Comment{\textit{sequence of length $T$}}
      \State $x^{l_s}_{\cdot \mid t} \leftarrow f(x^{l}(t),u^{l_s}_{\cdot \mid t},0)$  \Comment{\textit{rollout over horizon $T$}}
  \EndPFor \label{line:traj_sample1}
  \State $V_{\theta}^{l}(x^{l_s}_{t+T \mid t}) \leftarrow \underset{z}{\mathbb{E}} \left[ g^l_{\theta}(z,x^{l_s}_{t+T \mid t})\right]$
  \PFor{$i = 1,\dots,P$}
        \State \edit{$\text{Compute } 
  J_\text{sampled}^{l}\bigl( x^{l}(t), x^{l_s}_{\cdot \mid t}, u^{l_s}_{\cdot \mid t};
  \lambda_i, V_{\theta}^{l}(x^{l_s}_{t+T \mid t})\bigr)$} 
\Statex \edit{$ \text{using \eqref{eq:sampled_cost_function}}$}  \label{line:cost}
        \State \edit{$\text{Compute }\hat{u}^{\star l}_{\cdot \mid t, \lambda_i} \text{ using \eqref{eq:importance_sampling_1} }$} \label{line:importance}
        \State $\hat{x}^{\star l}_{\cdot\mid t,\lambda_i} \leftarrow f\bigl(x^{l}(t),\hat{u}^{\star l}_{\cdot \mid t,\lambda_i},0\bigr)$
  \EndPFor
    \State \edit{$\text{Select } \lambda^\star \text{ \text{ using \eqref{eq:control_selection} 
        }}$}\label{line:selection}
    \State return $\hat{u}^{\star l}_{t \mid t,\lambda^{\star}}$
\end{algorithmic}
\end{algorithm}
\section{Experiments}
To validate our approach, we conduct three experiments with progressively increasing system complexities. We start with a deterministic linear point-mass model to show that even for a deterministic linear system, SIT-LMPC still outperforms both LMPC and ABC-LMPC. Then, we move on to experiments with autonomous racing cars to showcase the advantages of SIT-LMPC when the system is stochastic, high-dimensional, nonlinear, and nonholonomic. To simulate stochasticity, truncated Gaussian noise is added to state observations and control inputs. Lastly, we validate our approach through real-world experiments with off-road autonomous racing on a $1/5$th scaled autonomous vehicle. Each experiment was conducted five times for each controller, with individual rollouts shown in light colors and the averaged performance shown in solid colors.

\subsection{Point-mass Navigation}

In this experiment, we consider the point mass system described in \cite{althoff2017commonroad} with $x = [p_x, p_y, v_x, v_y]^\mathrm{T}$ where $p_x$, $p_y$ are positions and $v_x$, $v_y$ are velocities in Cartesian coordinates. 

\begin{figure}[t]
  \centering
  \subfigure[]{\includegraphics[width=0.48\linewidth]{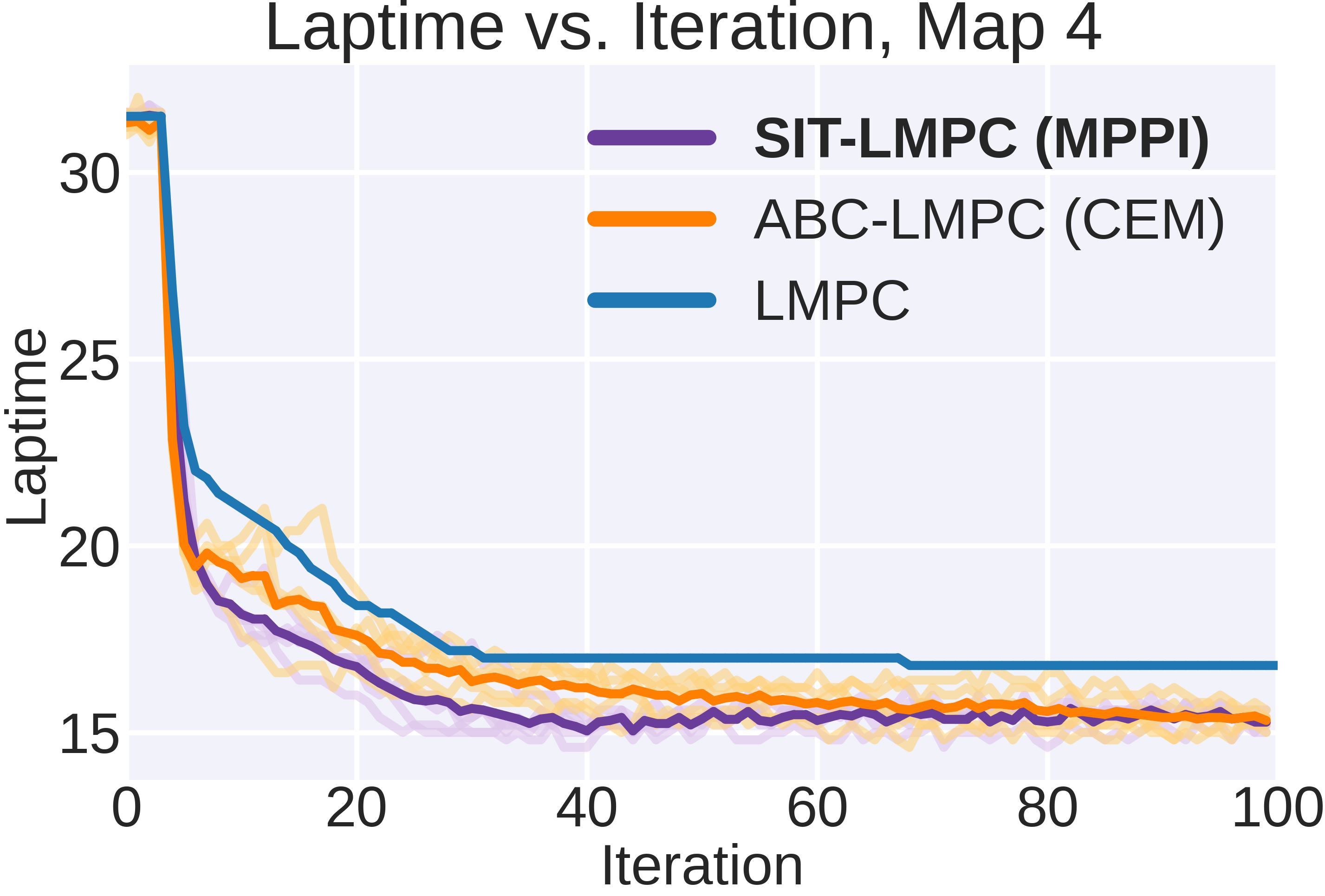}\label{fig:point_mass_laptime}}
  \hfill
  \subfigure[]{\includegraphics[width=0.48\linewidth]{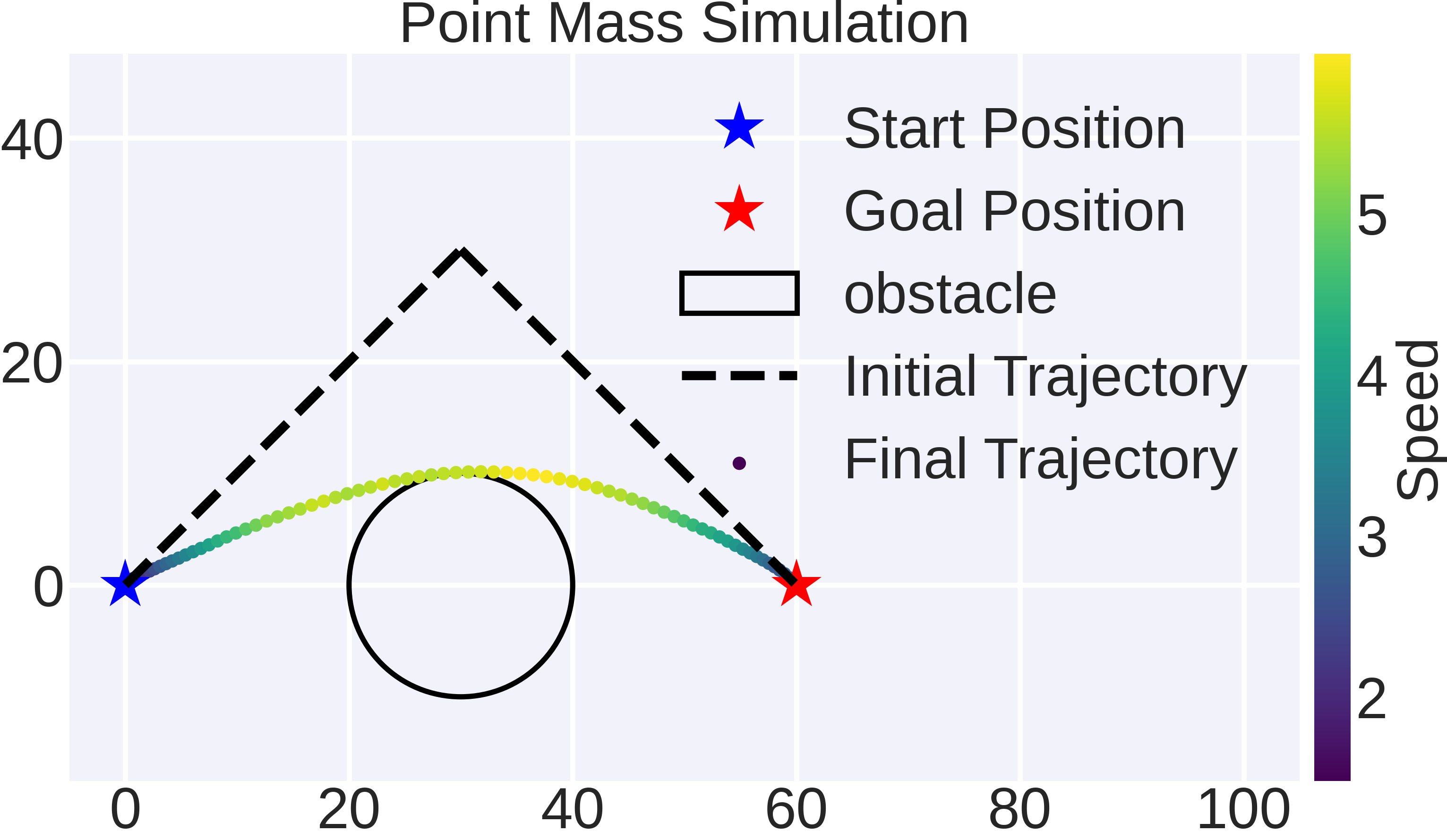}\label{fig:point_mass_layout}}
  \vspace{-1em}
  \caption{Point mass experiment. (a) Convergence of lap time over iterations. (b) Layout of the experiment.}
  \label{img/point_mass1}
  \vspace{-1.5em}
\end{figure}

The control objective is to steer the system from the initial state $x_0 = [0, 0, 0, 0]^\mathrm{T}$ to the target state $\mathcal{T} = \{[60, 0, 0, 0]^\mathrm{T}\}$ in minimum time while avoiding a circular obstacle centered at $p_{obs} = [30, 0]^\mathrm{T}$ of radius $10$ so that $\mathcal{X} = \{ [p_x, p_y]^\mathrm{T} \in \mathbb{R}^2 : \| [p_x, p_y]^\mathrm{T} - p_{obs}\|_2 > 10 \}$. The control input $u = [a_x, a_y]^\mathrm{T}$ is the acceleration in Cartesian coordinates and is assumed to be constrained within the set $\mathcal{U} = \{ u \in \mathbb{R}^2: -1 \leq a_x \leq 1,\; -1 \leq a_y \leq 1 \}$. The safe set is initialized with a hand-designed demonstration trajectory. The bottom part of Fig. \ref{img/point_mass1} shows the layout of this experiment. As shown in the top part of Fig. \ref{img/point_mass1}, all three methods can iteratively reduce the time to reach the target without crashing into the obstacle, with SIT-LMPC outperforming in terms of convergence rate. This experiment shows that even for a deterministic 2D linear system with non-linear non-convex constraints, SIT-LMPC outperforms prior methods. Even though the system is linear, we solve a non-convex optimization problem due to the obstacle-avoidance constraint. Hence, it is expected that the LMPC, which uses a gradient-based solver, converges to a local minimum, and the sampling-based methods manage to escape the local minimum to converge to a better solution.

\subsection{Vehicle Trajectory Optimization with Simulated Dynamics}
\label{section/dynamicST}
The benefits of SIT-LMPC are accentuated with nonlinear, stochastic, and nonholonomic systems. To show this, we compare SIT-LMPC with ABC-LMPC on an autonomous race car tasked with minimizing lap time while staying within the track boundaries. \edit{Autonomous racing involves nonlinear dynamics, safety-critical constraints, and real-time decision-making, making it a strong benchmark for robotics \cite{10924398}. Specifically, it requires a balance between safety and performance, as minimizing lap time demands pushing to the limits while remaining safe}. In this experiment, we use the dynamic single-track model from \cite{althoff2017commonroad}, with parameters of vehicle ID:1, expressed in Frenet coordinates. The state of the system is given by $x = [p_x, p_y, v, \delta, \Psi, \dot{\Psi}, \beta]^\mathrm{T}$, where $(p_x, p_y)$ is the position in cartesian coordinates, $v$ is the velocity in the $x-$direction, $\delta$ is the steering angle of the front wheels, $\Psi$ is the heading of the vehicle, $\dot{\Psi}$ is the yaw rate of the vehicle, and $\beta$ is the side-slip angle of the vehicle. The control input $u = [a, \delta_v]^\mathrm{T}$ is the acceleration in the longitudinal direction and the steering speed. To define our initial state $x_0$, the target set $\mathcal{T}$, and the set of admissible states $\mathcal{X}$, we express the vehicle's position in Frenet coordinates using the transformation detailed in \cite{micaelli1993trajectory}. The resulting pose in Frenet frame is $[s, e_y, e_\Psi]^\mathrm{T}$, where $s$ is the progress along the track, $e_y$ is the lateral displacement, and $e_\Psi$ is the yaw angle in vehicle frame. The initial state is the start line with $s = 0$ and the target set is $\mathcal{T} = \{x\in\mathbb{R}^7 : s \geq L_\mathrm{t}\}$, where $L_\mathrm{t}$ is the arc length of the track. The set of admissible states $\mathcal{X}$ is defined by $\{x \in \mathbb{R}^7 : -w \leq e_y \leq w\}$ where $w$ is the width of the track, assumed to be uniform. The control objective is to minimize the stage cost function $h\left(x, u \right) \triangleq \mathds{1}_{\mathcal{X} \setminus \mathcal{T}}(x)$, where $\mathds{1}_{\mathcal{X} \setminus \mathcal{T}}(\cdot)$ is the indicator function of the set $\mathcal{X} \setminus \mathcal{T}$. In other words, this stage cost incurs a cost of $1$ until the vehicle crosses the finish line to the target set $\mathcal{T}$.

\begin{figure*}[t]
  \centering
  \subfigure{\includegraphics[width=0.32\textwidth]{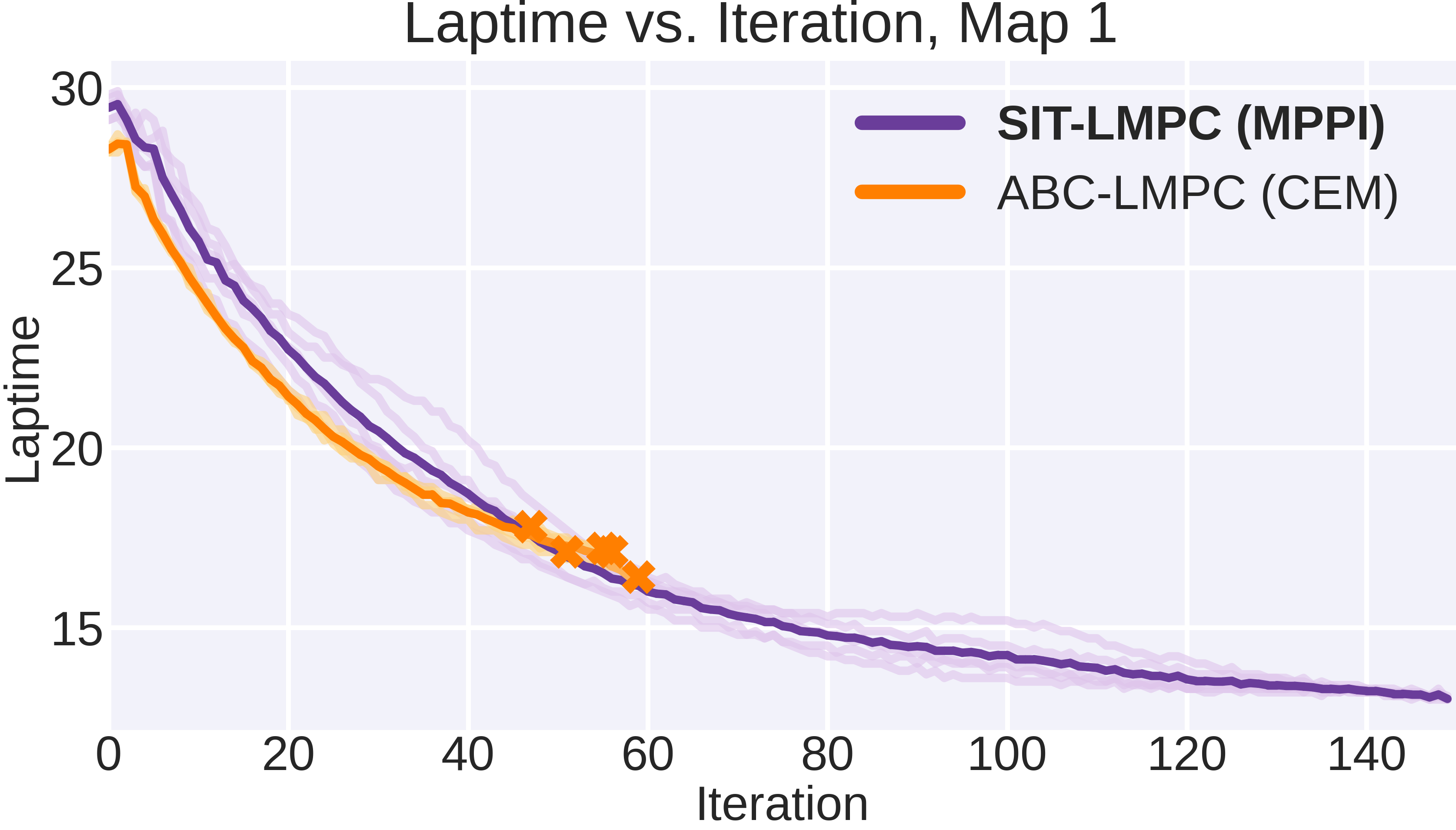}\label{fig:laptime55}}
  \hfill
  \subfigure{\includegraphics[width=0.32\textwidth]{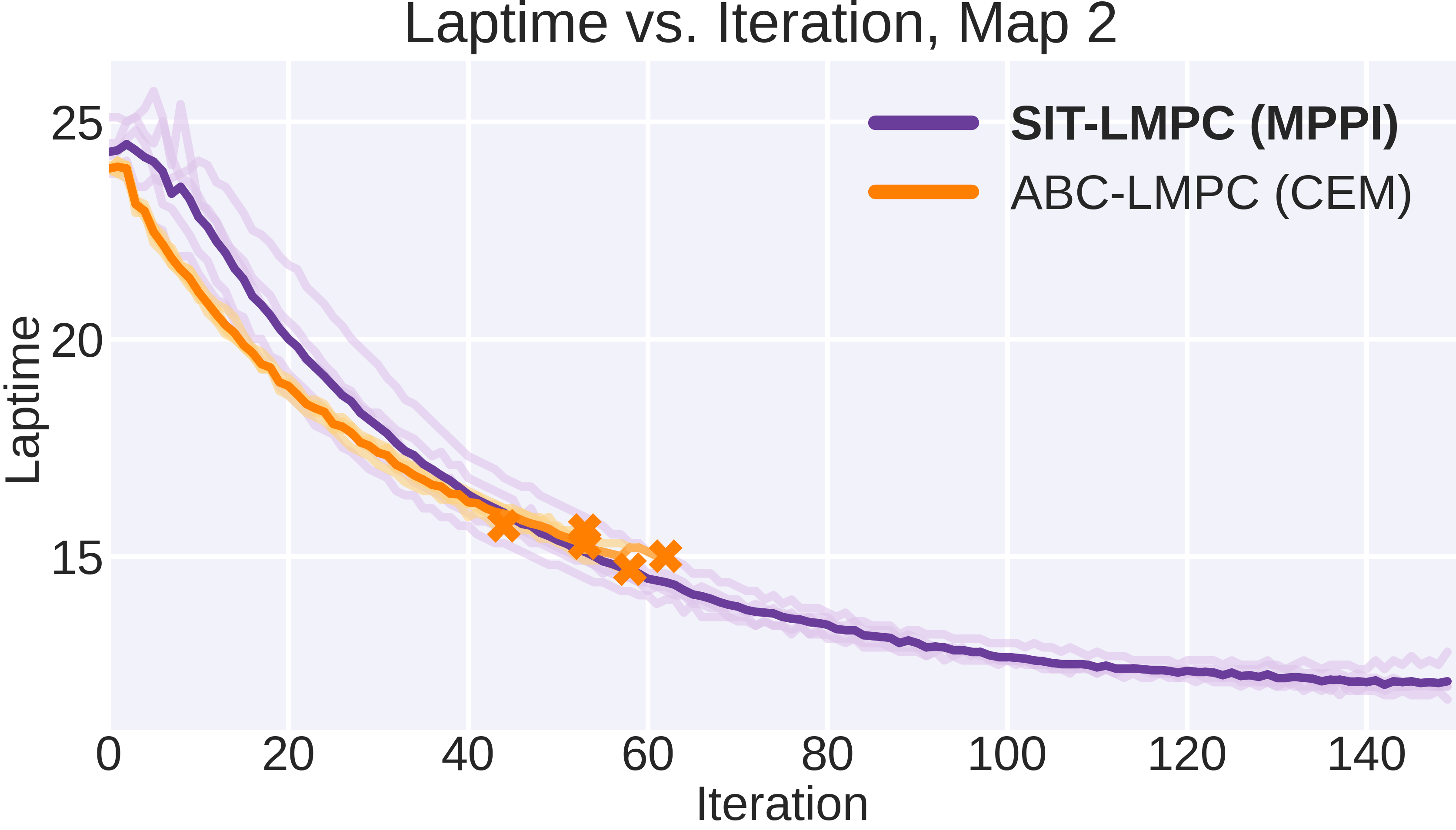}\label{fig:laptime58}}
  \hfill
  \subfigure{\includegraphics[width=0.32\textwidth]{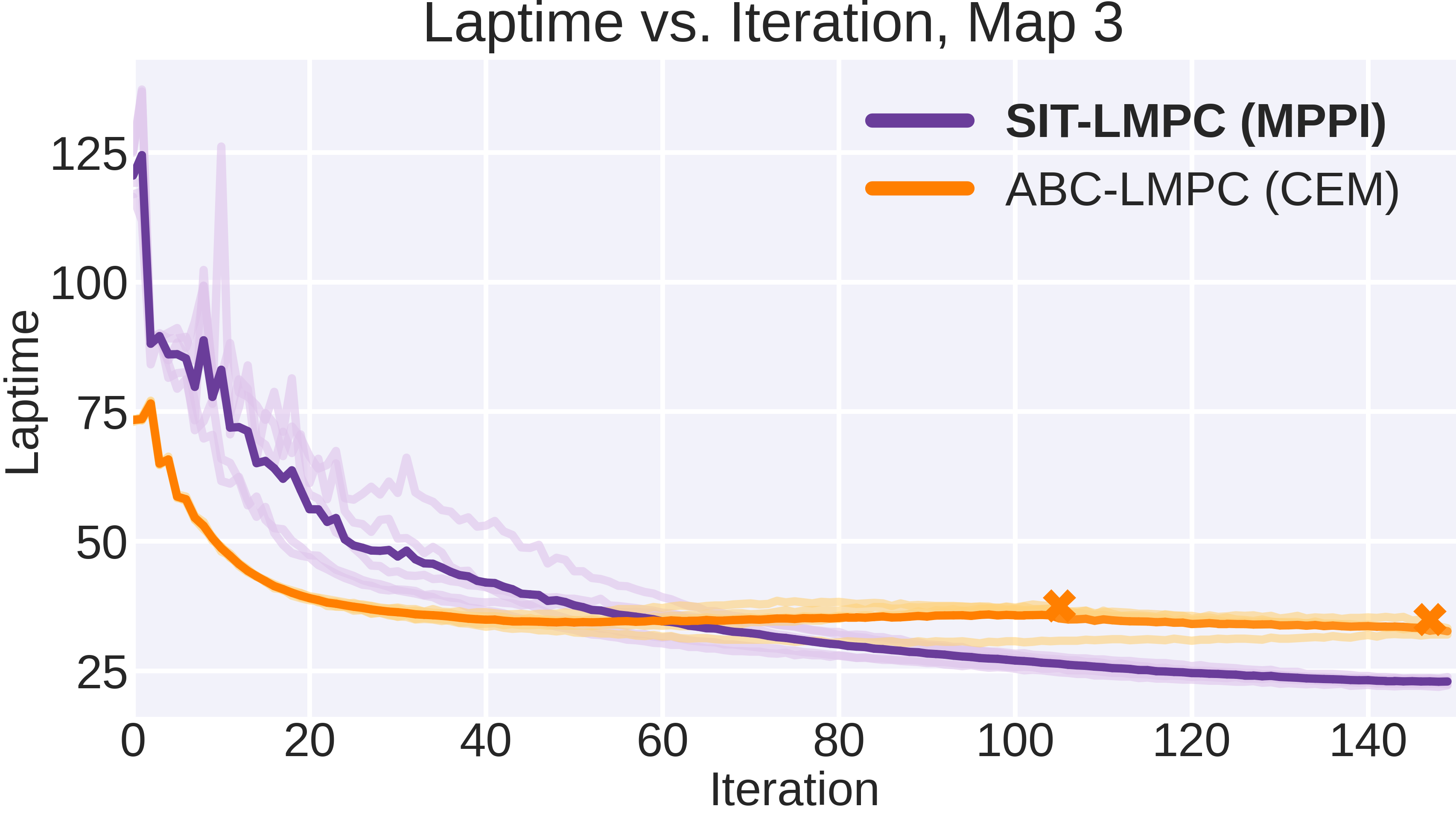}\label{fig:laptime60}}
  \vspace{-0.6em}
  \par
  \subfigure{\includegraphics[width=0.32\textwidth]{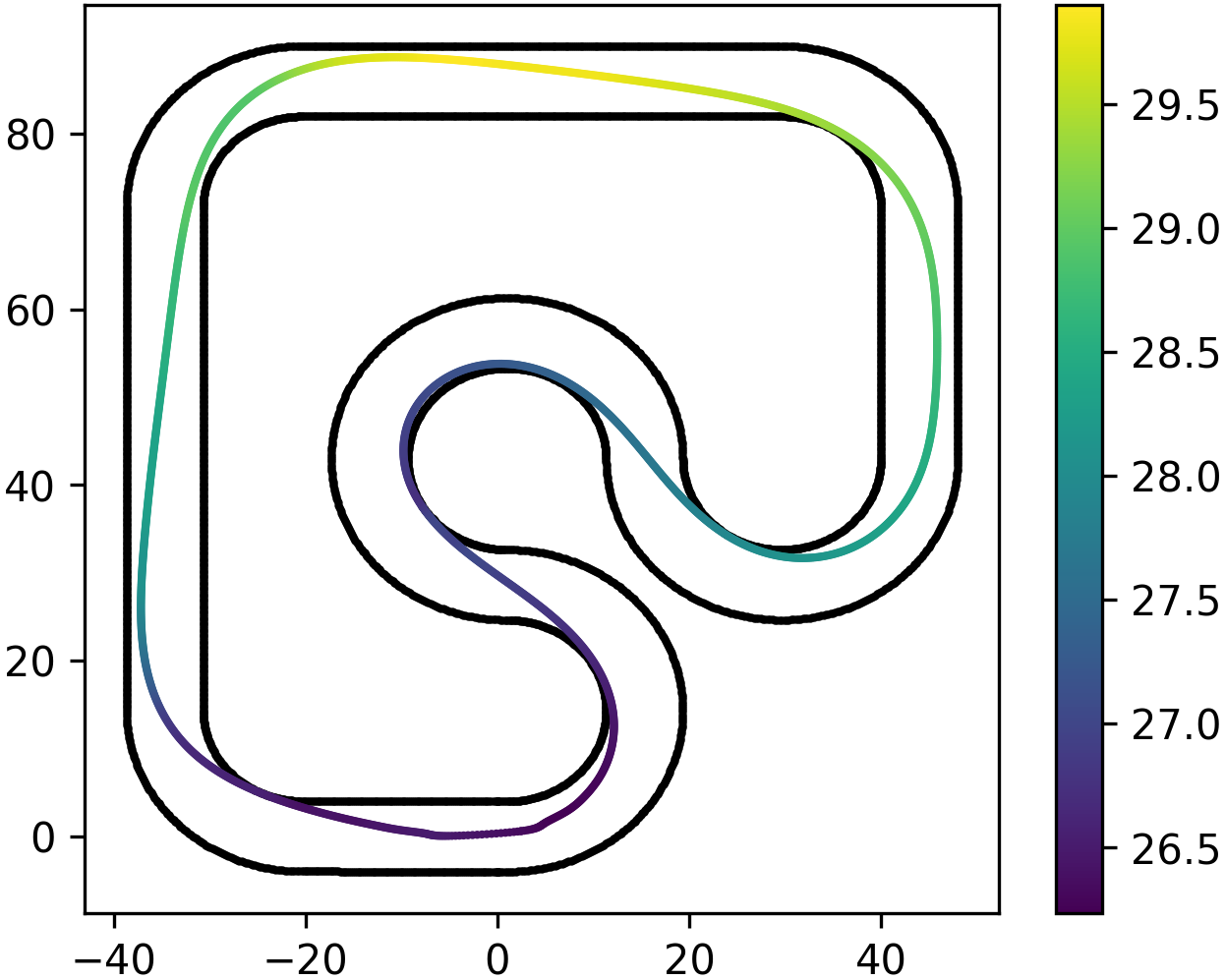}\label{fig:traj55}}
  \hfill
  \subfigure{\includegraphics[width=0.32\textwidth]{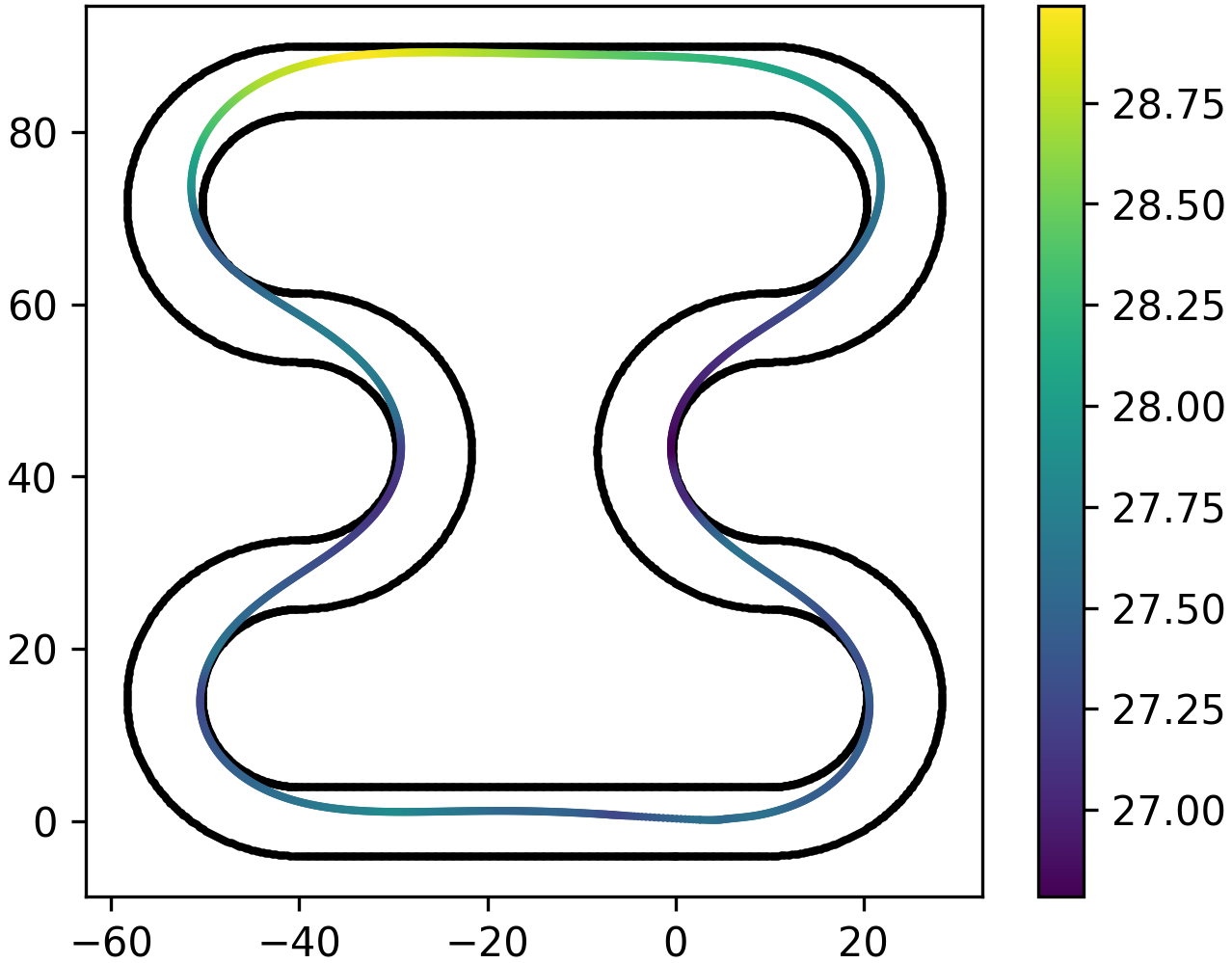}\label{fig:traj58}}
  \hfill
  \subfigure{\includegraphics[width=0.32\textwidth]{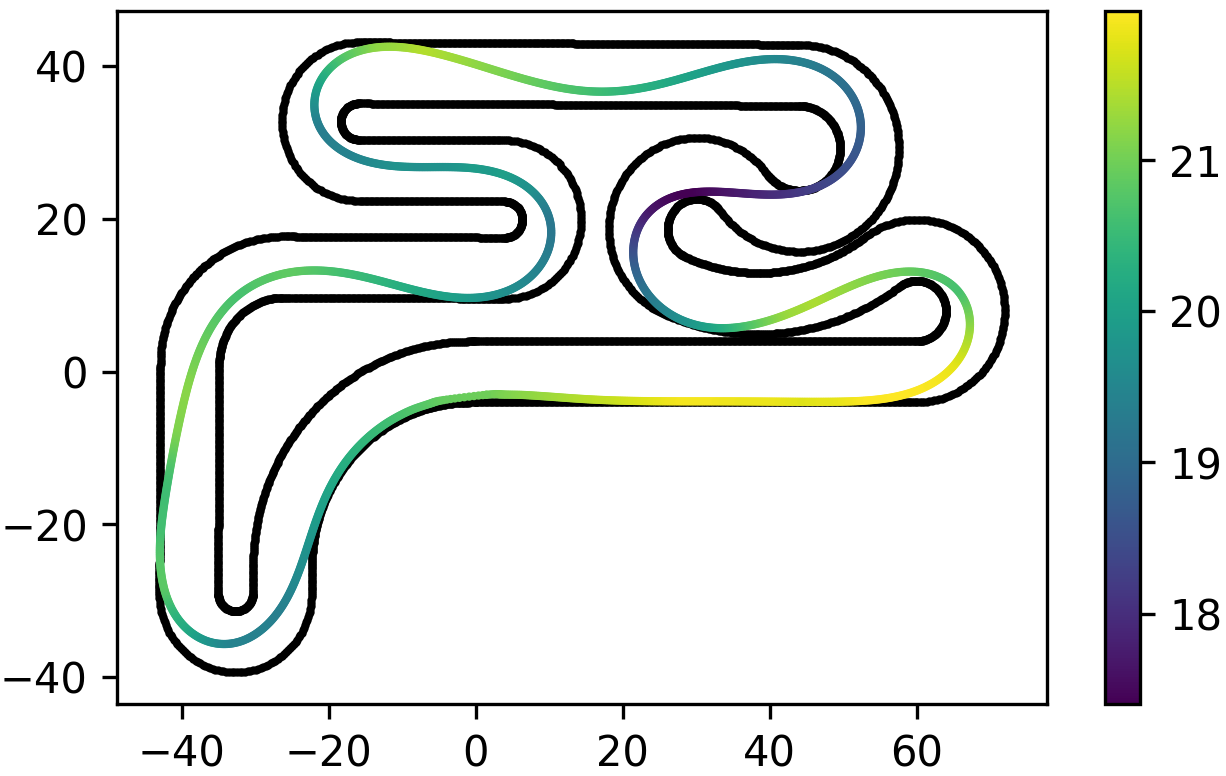}\label{fig:traj60}}
  \caption{Top: Convergence of lap time for three simulated experiments (five independent rollouts in light color and averages in dark color; $\times$ denotes out-of-track). Bottom: Fastest lap trajectories with velocity colorbar.}
  \label{img:racecar_laptime}
\end{figure*}

The top part of Fig. \ref{img:racecar_laptime} shows the lap time of the vehicle over successive iterations. ABC-LMPC frequently crashes halfway through the iterations, failing to satisfy the safe set and state constraints. We attribute this primarily to the selection of the single least-cost sample trajectory by CEM. Consequently, when this sample trajectory approaches a constraint boundary, system noise or perturbations can cause the vehicle to crash. We also tested running ABC-LMPC with multiple CEM iterations, but this did not yield notable performance improvements despite the additional computational cost. In contrast, SIT-LMPC converges to a lower lap time while ensuring safety throughout all $150$ iterations. Note that when LMPC failed to finish more than five iterations without crashing, as LMPC is \textit{not} designed for stochastic nonlinear systems.

\begin{figure}[!tb]
  \centering
  \subfigure[Ablations with CEM]{\includegraphics[width=0.49\linewidth]{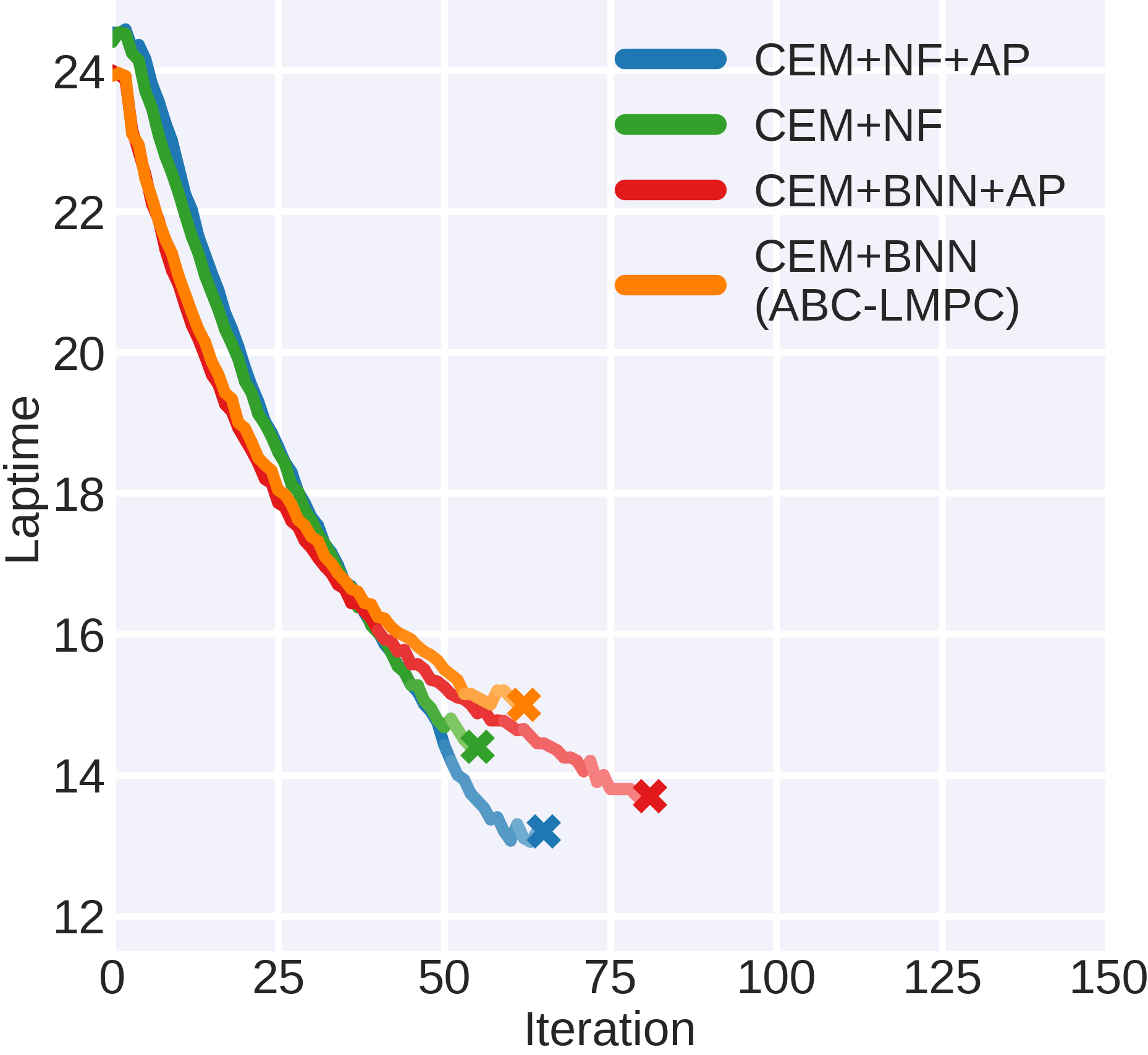}\label{img:ablation_cem}}
  \hfil
  \subfigure[Ablations with MPPI]{\includegraphics[width=0.49\linewidth]{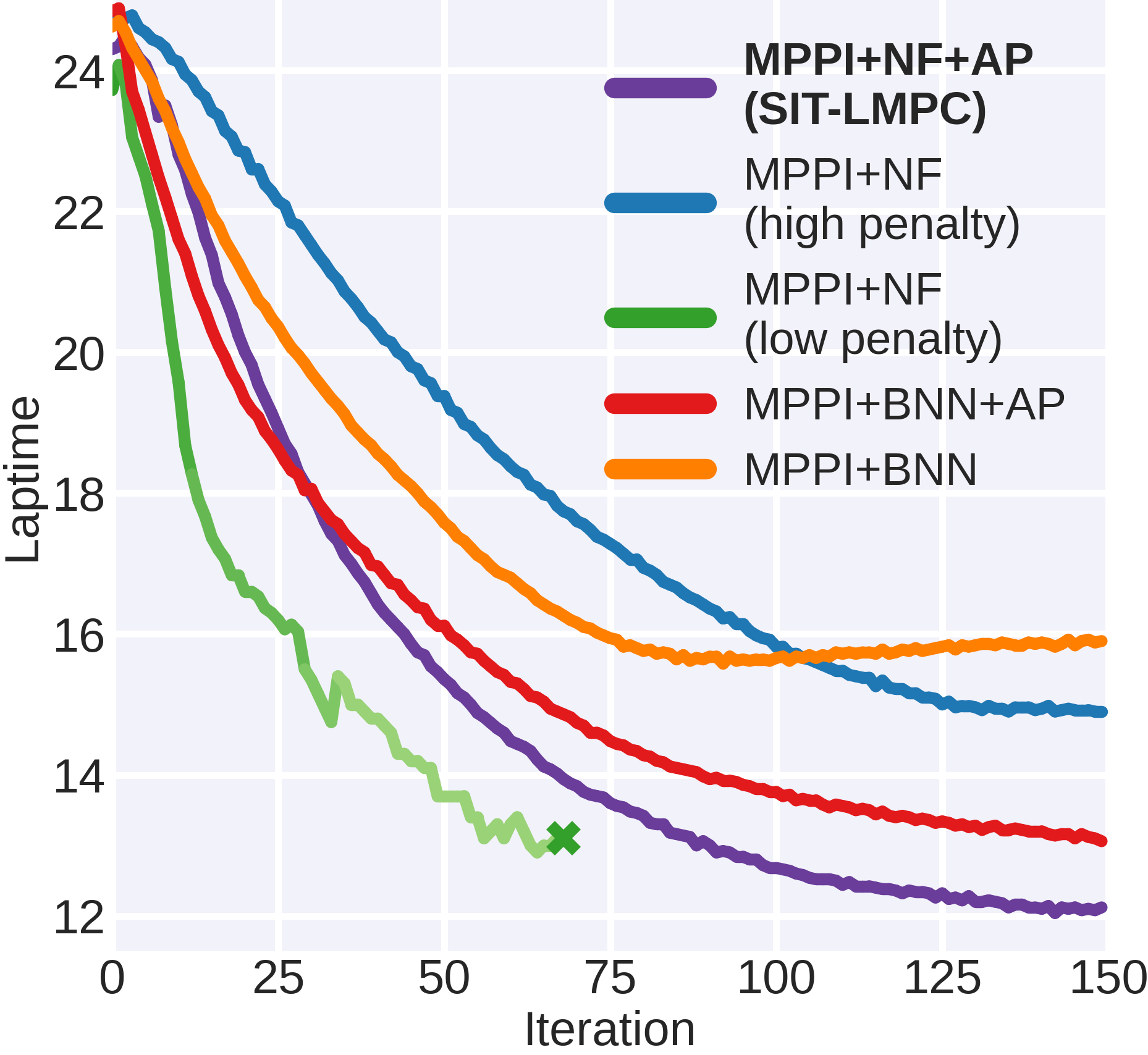}\label{img:ablation_mppi}}
  \caption{\edit{Ablation study of key SIT-LMPC components for a CEM controller (left) and an MPPI controller (right), comparing NF and BNN value function models, with and without the AP method. Plotted are averages of five rollouts with $\times$ denoting out-of-track.}}
  \label{img:ablation}
  \vspace{-1.5em}
\end{figure}

\textbf{Ablation Study:} Using the simulated racing environment, we 
\edit{perform ablation studies on the key components of SIT-LMPC. As shown in Fig. \ref{img:ablation_cem}, using CEM always yields infeasibility. We believe this is due to CEM's susceptibility to noise perturbation and mode collapse. For both CEM and MPPI, comparing the experiments using the NFs and the BNNs, the NF-based modeling yields better performance. To show the effect of the proposed AP method, we tested SIT-LMPC with a fixed penalty. Fig. \ref{img:ablation_mppi} shows that a fixed high penalty (blue) yields suboptimal performance, whereas a fixed low penalty (green) makes the system unsafe. We also implemented each combination with and without the AP method. Our results show that AP improves the learning process, even when used for ABC-LMPC (red in Fig. \ref{img:ablation_cem}), but is most effective as an integral part of the SIT-LMPC.}

\subsection{Real-world Experiment with 1/5th Scaled Vehicle}
\begin{figure}[t]
  \centering
  \subfigure[Experimental platform]{\includegraphics[width=0.48\linewidth]{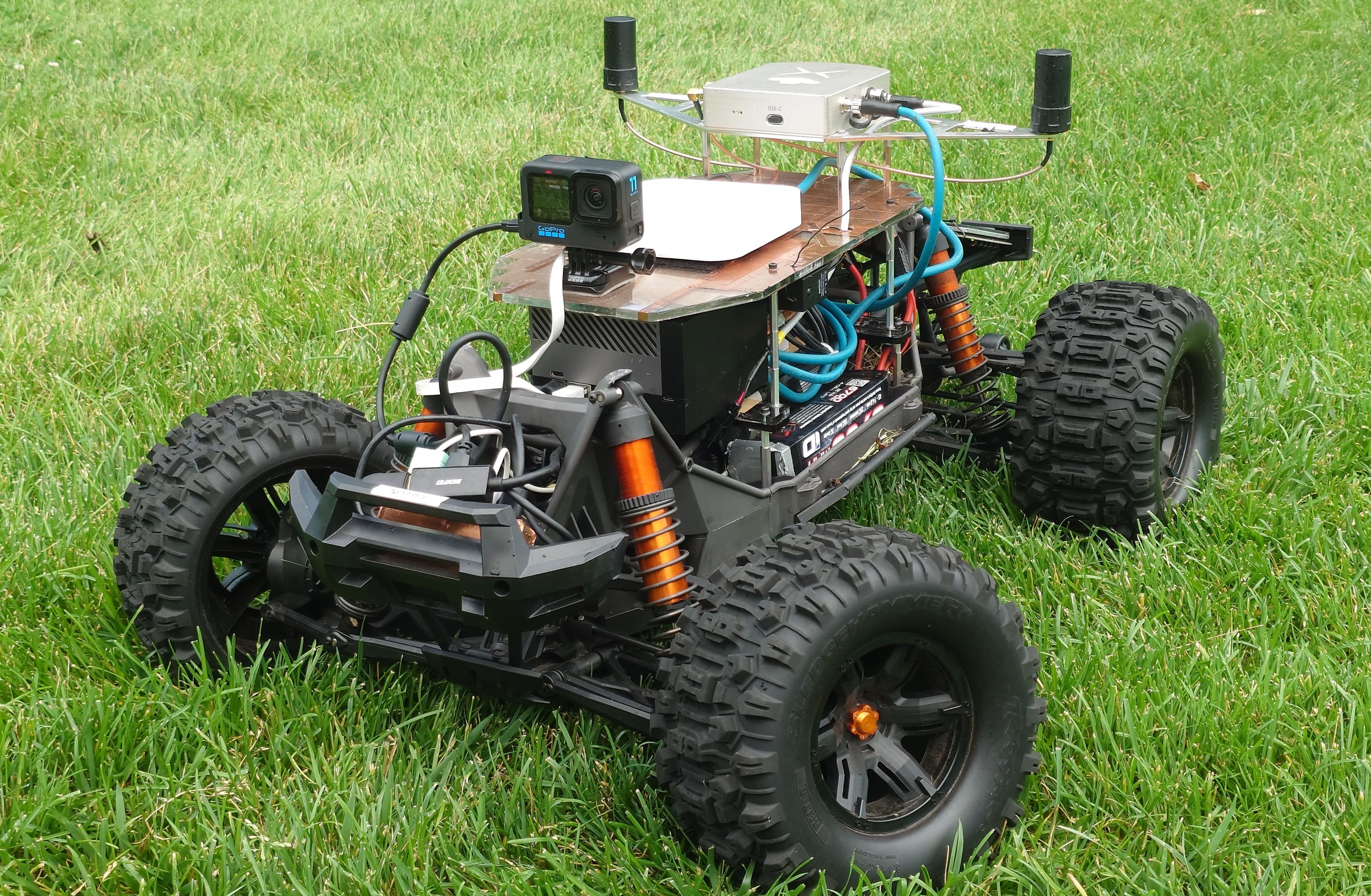}\label{fig:robot-platform}}
  \hfill
  \subfigure[Track layout]{\includegraphics[width=0.48\linewidth]{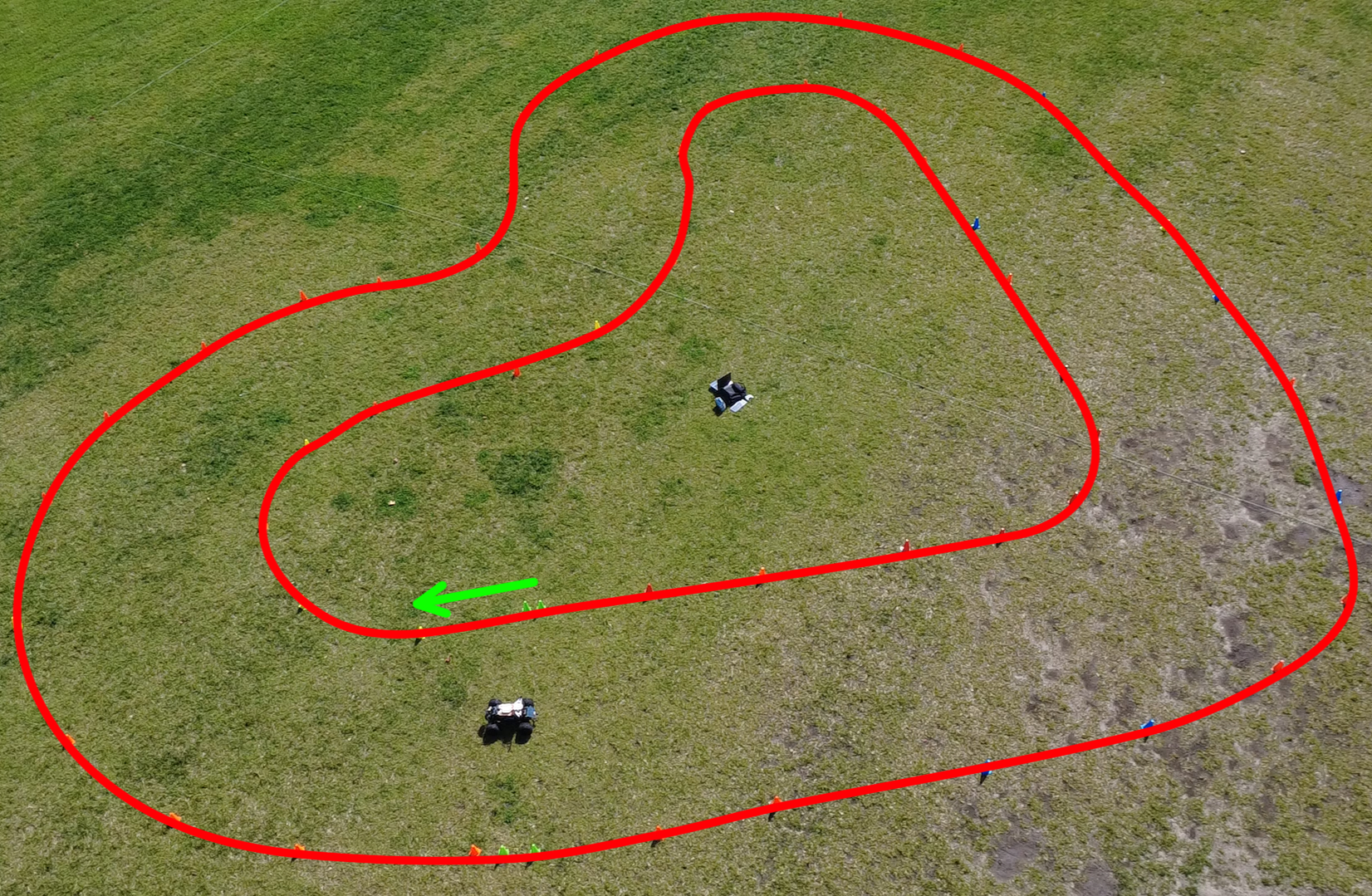}\label{fig:track-layout}}
  \vspace{0.5em}
  \par
  \subfigure[Laptime per iteration]{\includegraphics[width=0.48\linewidth]{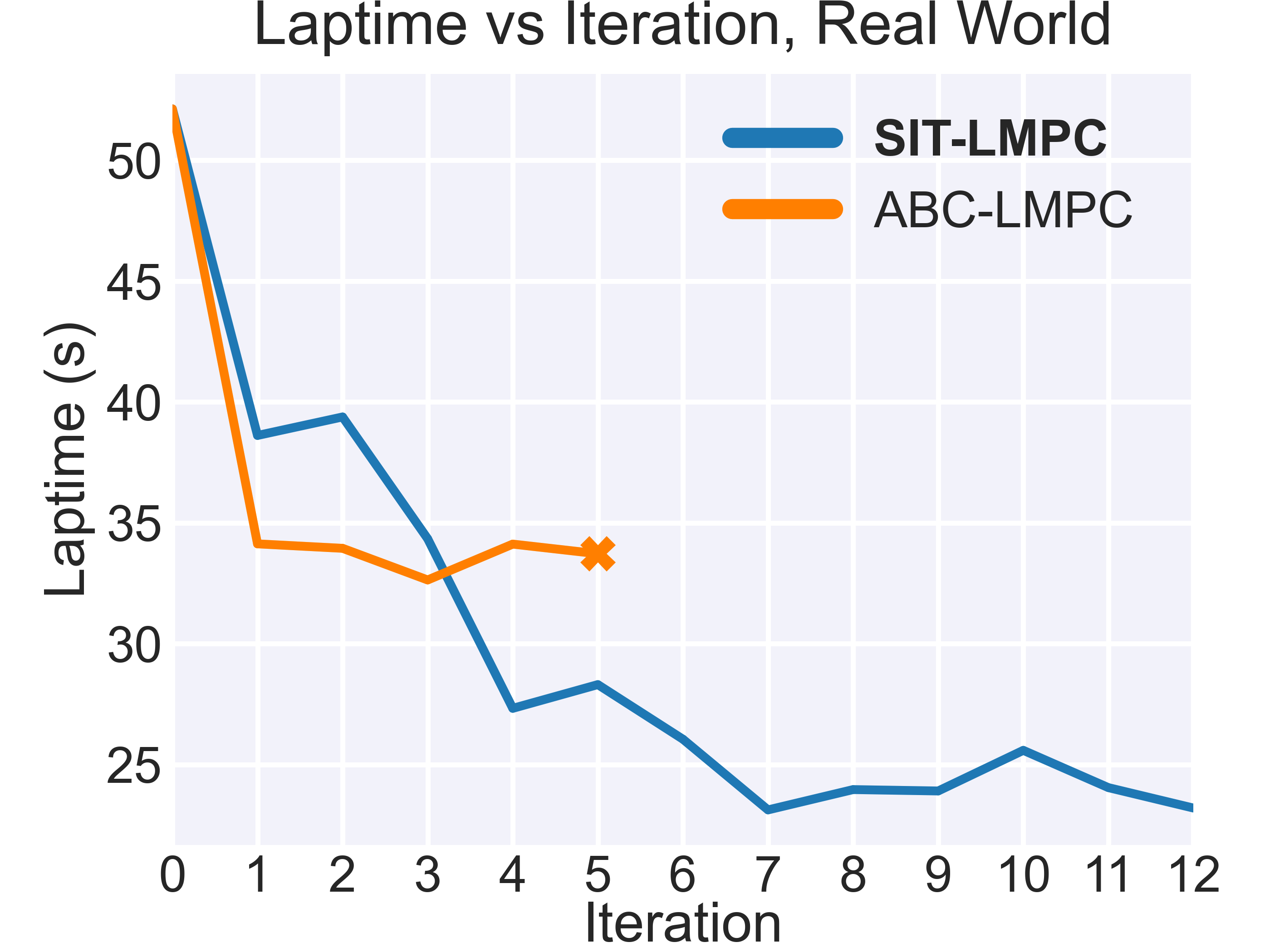}\label{fig:laptime-real}}
  \hfill
  \subfigure[Boundary violations]{\includegraphics[width=0.48\linewidth]{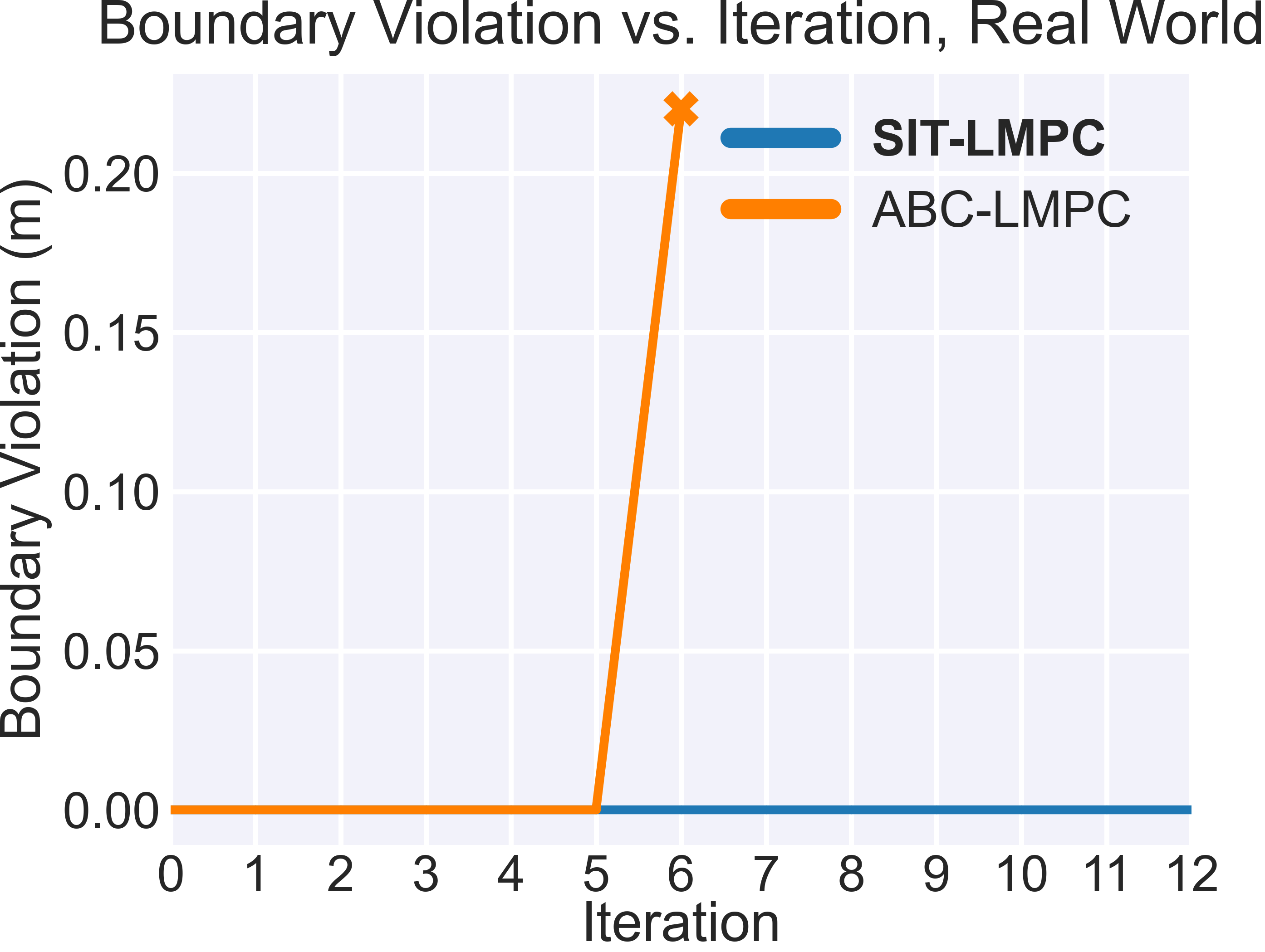}\label{fig:boundary-real}}
  \caption{Experimental setup and results with a real vehicle. (a) Platform. (b) Track. (c) Laptime per iteration. (d) Boundary violations.}
  \label{fig:real-results}
  \vspace{-2em}
\end{figure}

To show the effectiveness of our approach on real robotic systems, we deployed SIT-LMPC on a $1/5$th scale autonomous off-road race car shown in Fig. \ref{fig:robot-platform}, \edit{comparing with ABC-LMPC as a baseline}. The initial state $x_0$, the target set $\mathcal{T}$, the set of admissible states $\mathcal{X}$, and the stage cost $h\left(x, u \right)$ for this experiment are identical to those of \ref{section/dynamicST}. The localization is provided by a Fixposition Vision-RTK 2 GNSS system in an off-road grassy outdoor environment. The racetrack used for this experiment is shown in Fig. \ref{fig:track-layout}. The initial safe set is created using an MPPI controller tracking the centerline of the track at a reference speed of $2$ m/s with an initial lap time of $52.13$ seconds. For both the MPPI controller and SIT-LMPC, we use the same dynamic model as in section \ref{section/dynamicST}. The parameters of this model were empirically identified, and are not entirely accurate. Fig. \ref{fig:laptime-real} shows that SIT-LMPC improves lap time (shown in blue) over iterations. During these iterations, the average velocity increased by $75\%$ up to $3.5$ m/s, limited by the acceleration bound of the car. \edit{Comparatively, ABC-LMPC does not converge without crashing. SIT-LMPC remains safe while outperforming ABC-LMPC by achieving a lap time $31.43\%$ faster. This shows that SIT-LMPC improves safety and performance for a real-world stochastic system with model mismatch, imperfect localization, and control noise from the off-road environment.} 

\section{Conclusion} 
This paper presented SIT-LMPC, an extension of the LMPC formulation to stochastic nonlinear systems. The resulting constrained receding-horizon stochastic optimization problem was solved by developing a constrained information-theoretic MPC algorithm using exterior penalty functions with online adaptive penalty parameters. The proposed method does not rely on prior assumptions about system dynamics or state constraints and is fully parallelizable by construction. Additionally, learning the value function using normalizing flows allowed for richer uncertainty modeling and yielded better performance compared to BNNs. Simulations and real-world experiments demonstrated that SIT-LMPC generates safer and higher-performance trajectories than LMPC and ABC-LMPC. Future research \edit{ will focus on applying SIT-LMPC to a wider range of robotic platforms and addressing systems with unknown dynamics.}

\bibliographystyle{IEEEtran}
\bibliography{IEEEabrv,bibs/references}

\end{document}